\documentclass[times,final]{elsarticle}

\usepackage{framed,multirow}

\usepackage{amssymb}
\usepackage{latexsym}
\usepackage{bm}
\usepackage{amsmath}
\usepackage[labelfont=bf,textfont=md]{caption}
\usepackage{graphicx}
\usepackage[utf8]{inputenc}
\usepackage{tikz}
\usepackage[ruled,vlined]{algorithm2e}
\usepackage{algorithmic}
\usepackage[utf8]{inputenc}
\usepackage[english]{babel}
\usepackage{amsthm}
\usepackage{amsmath}

\usepackage[lofdepth,lotdepth]{subfig}

\graphicspath{ {./images/} }

\usetikzlibrary{shapes.geometric, arrows}
\tikzstyle{startstop} = [rectangle, rounded corners, minimum width=3cm, minimum height=1cm,text centered, draw=black, fill=red!30]
\tikzstyle{io} = [trapezium, trapezium left angle=70, trapezium right angle=110, minimum width=3cm, minimum height=1cm, text centered, draw=black, fill=blue!30]
\tikzstyle{process} = [rectangle, minimum width=3cm, minimum height=1cm, text centered, draw=black, fill=orange!30]
\tikzstyle{decision} = [diamond, minimum width=3cm, minimum height=1cm, text centered, draw=black, fill=green!30]
\tikzstyle{arrow} = [thick,->,>=stealth]

\theoremstyle{definition}
\newtheorem{definition}{Definition}[section]
\theoremstyle{remark}

\newcommand{\indep}{\perp \!\!\! \perp}
\usepackage{url}
\usepackage{xcolor}
\definecolor{newcolor}{rgb}{.8,.349,.1}

\begin{document}

\begin{frontmatter}

\title{RMFGP: Rotated Multi-fidelity Gaussian process with Dimension Reduction for High-dimensional Uncertainty Quantification\tnoteref{tnote1}}%


\author[1]{Jiahao {Zhang}\corref{cor1}}

\author[1]{Shiqi {Zhang}\corref{cor1}}

\author[1,2]{Guang {Lin}\corref{cor2}}

\cortext[cor1]{These authors contributed equally.}

\cortext[cor2]{Corresponding author. 
E-mail: guanglin@purdue.edu.}

\address[1]{Department of Mathematics, Purdue University, West Lafayette, IN 47906, USA}
\address[2]{School of Mechanical Engineering, Department of Statistics (Courtesy), Department of Earth, Atmospheric, and Planetary Sciences (Courtesy), Purdue University, West Lafayette, IN 47907, USA}



\begin{abstract}
Multi-fidelity modelling arises in many situations in computational science and engineering world. It enables accurate inference even when only a small set of accurate data is available. Those data often come from a high-fidelity model, which is computationally expensive. By combining the realizations of the high-fidelity model with one or more low-fidelity models, the multi-fidelity method can make accurate predictions of quantities of interest. This paper proposes a new dimension reduction framework based on rotated multi-fidelity Gaussian process regression and a Bayesian active learning scheme when the available precise observations are insufficient. By drawing samples from the trained rotated multi-fidelity model, the so-called supervised dimension reduction problems can be solved following the idea of the sliced average variance estimation (SAVE) method combined with a Gaussian process regression dimension reduction technique. This general framework we develop can effectively solve high-dimensional problems while the data are insufficient for applying traditional dimension reduction methods. Moreover, a more accurate surrogate Gaussian process model of the original problem can be obtained based on our trained model. The effectiveness of the proposed rotated multi-fidelity Gaussian process(RMFGP) model is demonstrated in four numerical examples. The results show that our method has better performance in all cases and uncertainty propagation analysis is performed for last two cases involving stochastic partial differential equations. 
\end{abstract}

\end{frontmatter}


\section{Introduction}
Many models of scientific computing and engineering are very expensive to evaluate, and yet the number of points needed to explore the entire area can be prohibitive, especially in high dimensional space, which is the famous \emph{curse of dimensionality problem}. The model complexity and computational cost both increase dramatically in this situation. At this point, dimension reduction techniques come to our rescue by discovering and employing the low-dimensional structure in the problem itself. A brief description about dimension reduction is that if the conditional distribution of the quantities of interest $Y$ given inputs $X$ depends on $X$ only through a matrix $\beta$ in the form of $\beta^TX$. Then the so-called dimension reduction space is spanned by the column of matrix $\beta$. The central subspace is the smallest such dimension reduction space. The details can be found in \cite{dimensionreduction}. There are various methods on finding such central space. For example, the principle component analysis (PCA) in \cite{pca} is the most famous unsupervised dimension reduction method. Many other techniques are developed based on different constructions to deal with different tasks. In this paper, we consider the supervised dimension reduction problems, in which the response $Y$ is often a scalar. Previous works relating to this type of problems can be found in \cite{acs, cr, SIR, InReUQ, save, GPbDR}.

The sliced inverse regression (SIR) proposed in \cite{SIR} and the sliced average variance estimation (SAVE) proposed in \cite{save} type of methods are very popular among all supervised dimension reduction techniques. Interested readers may refer to \cite{dimensionreduction} for a more comprehensive review of this topic. However those methods may not work well in the situation where available data come from models of different fidelity levels, thus high-fidelity data are insufficient and expensive to obtain. The authors in \cite{BIRsd} proposed a Bayesian approach to compute the conditional distribution $\pi(\bold{x}|y)$ of the predictors $\bold{x}$ given the response variable $y$ in order to perform dimension reduction. The likelihood function in their method is obtained by using the Gaussian process regression model. Then $\pi(\textbf{x}|y)$ can be computed by Monte Carlo sampling. However, the Gaussian process model may not work well when the accurate observations are vary rare or the observations come from different fidelity levels thus affecting the performance of the dimension reduction. In this paper a rotated multi-fidelity Gaussian process is combined with SAVE type of methods to perform dimension reduction in this situation and a more accurate surrogate model can be obtained afterwards.
 
 Multi-fidelity modelling aims at combining the information in the low-fidelity models that can be inaccurate but inexpensive with that in the high-fidelity model which is computationally demanding. Peherstorfer et al. \cite{Pesurvey} give a complete review of multi-fidelity modelling methods with a focus on the application of uncertainty propagation, statistical inference and optimization. Among all different multi-fidelity modelling approaches, the one based on the Gaussian process regression \cite{CEGP} has been frequently used. The auto-regressive scheme put forth by Kennedy and O'Hagan \cite{autoregressive} exploits the linear correlation between the high-fidelity and low-fidelity models to improve the prediction accuracy. An efficient recursive implementation by Le Gratiet and Garnier \cite{Le} considerably reduces the complexity of the original auto-regressive scheme. However, there exists no simple linear correlations between different fidelity models or the linear correlations only exist in a specific range of inputs in many practical problems of interest. In those cases, the auto-regressive scheme tends to ignore the low-fidelity data and may return inaccurate predictions. To address this type of problems, Perdikaris et al. \cite{NARGP} proposed a nonlinear information fusion algorithm based on the auto-regressive scheme and the idea of Le Gratiet and Garnier \cite{Le}. It not only allows to learn complex nonlinear correlations between different fidelity models but also works in situations where only linear correlation exists. Since the high-fidelity model are usually computationally demanding, an active learning scheme is often needed for the problem with limited budget. The previous works relate to active learning can be found in \cite{AL1, AL3, AL2, AL4}. For simplicity, only two fidelity levels are considered in this paper but the method can be easily extended to the cases more than two fidelity levels. 
 
 In the first step of our method, a rotation matrix can be obtained based on the training data from low-fidelity model by using SAVE method and all training data are rotated by this matrix. This step aims at extracting some meaningful information from the low-fidelity data before feeding training data into the proposed multi-fidelity model and is proved to be useful in the numerical examples. Now, the nonlinear information fusion algorithm by Perdikaris et al. \cite{NARGP} is used as the building block of our multi-fidelity model and the predictions on pre-set test points can be obtained. With those predictions, SAVE method is employed again to find a rotation matrix. At this point, our trained multi-fidelity model can be less accurate due to lack of high quality data or it can be improved by expanding the realizations of high-fidelity model if the budget permits. This can be achieved by introducing an active learning scheme based on the problem setting. Once the stop criterion in the active learning scheme is reached, the rotation matrix in this step can be determined.
 The final dimensional reduction matrix and a surrogate model for the original problem will be deduced differently depending on an user defined parameter $flag$. If $flag = 0$, then the data for the final surrogate model are simply rotated by the rotation matrix from the last step and fed into a Gaussian process model. If $flag = 1$, the number of sufficient dimensions can be computed by Bayesian information criterion(BIC) method based on the trained rotated multi-fidelity model. The final dimensional reduction matrix is computed by combining the previous model and a Gaussian process dimension reduction technique with similar idea in \cite{GPbDR}. However the method in \cite{GPbDR} cannot be directly applied to our problem because the amount of high-fidelity data is not enough for the optimization process. The previous RMFGP model first reduced the number of original dimensions $p$ to $s$. Then a two-step Gaussian process optimization process is performed to find a reduction matrix to further reduce the number of dimensions to $d$ which is pre-computed by BIC. In the end, a new surrogate model for the original supervise dimension reduction problem, i.e. a Gaussian process model can be constructed using the data pre-processed by the final dimension reduction matrix.
 
 In this way, a rotated multi-fidelity Gaussian process model(RMFGP) is obtained and the inference process can be performed in two ways depending on the needs of the user.

Our objective of this paper:
\begin{enumerate}
    \item Find the intrinsic dimension in supervised dimension reduction problems with relatively small data set.
    
    \item Build an accurate surrogate model for high-dimensional problems with limited high-fidelity data.
\end{enumerate}

Our contribution in this paper:
\begin{enumerate}
\item A rotated multi-fidelity Gaussian process(RMFGP) model is proposed for high-dimensional problems with insufficient training data and the general workflow(Figure $1$) is developed.

\item The RMFGP model is combined with a two-step Gaussian process optimization process(Algorithm $3$) to find the final dimension reduction matrix. The BIC method(Section $2.4$) is applied to determine the reduced dimension.

\item An active learning scheme(Section $2.6$) in multi-fidelity GP is performed to improve the prediction accuracy of the RMFGP model. In the situation when high-fidelity data are rare and expensive to obtain, this is crucial for model performance.

\item Depending on the needs of the user, an accurate surrogate Gaussian process model for the original problem can be built based on the proposed dimension reduction method (Algorithm $4$). If the parameter $flag=0$, then the inputs are simply rotated by the matrix deduced from RMFGP. Otherwise the parameter $flag=1$, the inputs are projected onto a low-dimensional space by the final dimension reduction matrix. The model performance is illustrated in four numerical examples.

\end{enumerate}

The paper is organized as follows. In Section 2, we give a brief introduction to the famous Gaussian process regression(GPR) and the multi-fidelity GPR model with nonlinear auto-regressive scheme. Then the SAVE dimension reduction method and Bayesian active learning scheme are briefly reviewed. In Section 3, our algorithm is proposed and in Section 4, four different numerical examples are presented to illustrate our methods. The uncertainty propagation analysis in two stochastic PDE examples are also conducted. We summarize our findings and provide some discussions in Section 5.
\begin{figure}
\centering
\includegraphics[scale=0.35]{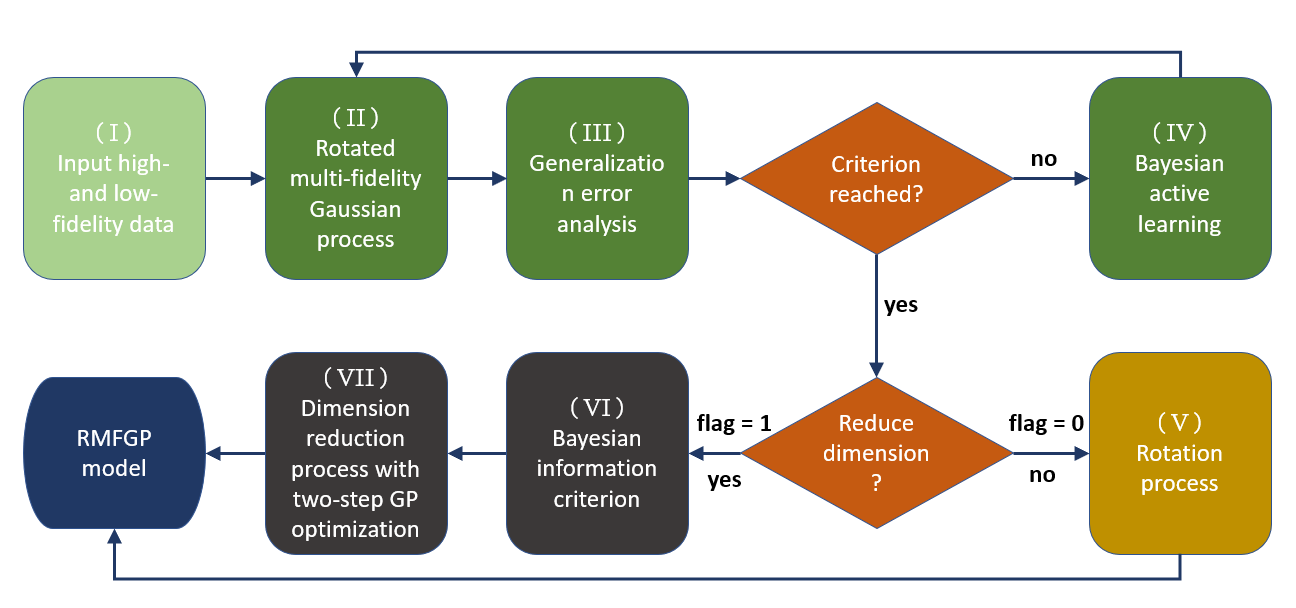}
\caption{\textbf{A general framework for solving high-dimensional problems with insufficient data - an overview of the workflow.} In this work, we first propose a rotated multi-fidelity Gaussian process model, in which the training data are rotated using the matrix computed by SAVE method before feeding to multi-fidelity GP model. Then, a generalization error analysis is performed. If the criterion is not fulfilled, a Bayesian active learning scheme is applied to add more high-fidelity data to training data set. Otherwise we obtain our RMFGP model. This model can be used to generate samples for computing a rotation matrix by SAVE method. If reduction parameter $flag=0$, a final GP model is built with input data rotated by this rotation matrix. If reduction parameter  $flag=1$, the BIC method is used to compute the final reduced dimension and then a two-step GP dimension reduction process is conducted to find the final reduction matrix. In this way, a GP model can be built with input data which are first projected onto a low-dimensional space by the final reduction matrix.}
\label{fig:flowchart}
\end{figure}

\section{Methodology}
Let $f: \mathcal{R}^p \rightarrow \mathcal{R}$ be a multivariate scalar function of $\bold{x}$ with input dimension $p>> 1$. Assume $f$ can be measured by a series of physical experiments with different accuracy or models with different computational cost. Furthermore, we allow for noisy measurements. So the observations have different levels of fidelity $t = 1,\cdots,r$. At each fidelity level $t$, the set of inputs is denoted by $D_t$. The set $D_r$ is the highest level and the corresponding response function $f_r$ is referred to as high-fidelity model. In this work, suppose $f_r$ has or can be approximated by the form,
$$
f_r(\bold{x}) \approx g(\bold{A}\bold{x})
$$
where the matrix $A$ projects $\bold{x}$ to a low dimensional subspace.

The goal of this paper is to construct an accurate model to determine the dimensional reduction matrix $A$ and a surrogate for the low dimensional map $g$. This is a supervised dimension reduction problem and we assume the observation set $D_r$ is relatively small and directly applying SAVE type of methods can perform poorly due to the lack of information in a limited high-fidelity data set. The first step is to construct an accurate rotated multi-fidelity model for the original high dimensional problem. The building blocks are the famous Gaussian process and the so-called nonlinear information fusion algorithm in \cite{NARGP}.


\subsection{Gaussian process regression framework}
The introduction in this section is based on \cite{CEGP}.  Interested readers can find a more thorough discussion in this book. Suppose there is an unknown mapping $z$:
$$y = z(\mathbf{x})$$ where $\mathbf{x} \in \mathcal{R}^p$ and $p$ is the input dimension.

The observations set can be denoted by $D = \{\mathbf{x}_i, y_i\}_{i=1}^n = (\mathbf{X}, \mathbf{y})$. In the Gaussian process framework, $z(\mathbf{x})$ is assumed to be a zero mean GP, i.e. $z \sim \mathcal{GP}(\mathbf{z}|\mathbf{0}, k(\mathbf{x}, \mathbf{x}';\mathbf{\theta})) $, where $k$ is an appropriate kernel function with a set of hyper-parameters $\mathbf{\theta}$. This assumption essentially reflects our prior belief about the function $z$.

If we assume a Gaussian likelihood, the optimal hyper-parameters in the kernel can be found by maximizing the marginal log-likelihood of the model,
$$
\mbox{log} p(\mathbf{y} | \mathbf{x}, \theta) = -\frac{1}{2} \mbox{log} |\mathbf{K}| - 
\frac{1}{2} \mathbf{y}^T \mathbf{K}^{-1} \mathbf{y} - \frac{n}{2} \mbox{log} 2\pi
$$
where $\mathbf{K} = (K_{ij})_{i,j = 1}^n$ and $K_{ij} = k(\mathbf{x}_i,\mathbf{x}_j;\theta)$.

The posterior distribution is thus tractable and the prediction for a new output $z_*$ at a new input $\mathbf{x}_*$ is given as 
$$
p(z_* | \mathbf{y}, \mathbf{X}, \mathbf{x}_*) = \mathcal{N}(z_*|\mu_*(\mathbf{x}_*), \sigma^2_*(\mathbf{x}_*))
$$

\begin{equation}
\mu_*(\mathbf{x}_*) = \mathbf{k}_{*n} \mathbf{K}^{-1} \mathbf{y}
\end{equation}

\begin{equation}
\sigma^2_*(\mathbf{x}_*) = \mathbf{k}_{**} - \mathbf{k}_{*n} \mathbf{K}^{-1} \mathbf{k}_{*n}^T
\end{equation}
where $\mathbf{k}_{**} = k(\mathbf{x}_*, \mathbf{x}_*)$ and $\mathbf{k}_{*n}  = [k(\mathbf{x}_*, \mathbf{x}_1), \cdots, k(\mathbf{x}_*, \mathbf{x}_n)]$. The posterior mean $\mu_*(\mathbf{x}_*)$ is the output of the model and the posterior variance quantified the uncertainties of the model about the predictions.

\subsection{Multi-fidelity Gaussian process with linear auto-regressive scheme}
To this end, suppose the data have $r$ levels of fidelity. At each level $t$, the output $y_t(\mathbf{x}_t)$ corresponding to each input $\mathbf{x}_t$ can be modeled by a Gaussian processes $Z_t(\mathbf{x}), t = 1, \cdots, r$. Then, the linear auto-regressive scheme is
$$ Z_t(\mathbf{x}) = \rho Z_{t-1}(\mathbf{x}) + \delta_t(\mathbf{x}), \hspace{2mm} t = 2,\cdots, r$$
where $\rho$ is the correlation coefficient between level $t-1$ and level $t$, $\delta_t(\mathbf{x})$ is a Gaussian process with mean $\mu_{\delta_t}$ and covariance function $k_t$. This construction implies the Markov property according to Kennedy and O'Hagan \cite{autoregressive}, which means there are nothing more about $Z_t(\mathbf{x})$ can be learned from other model $Z_{t-1}(\mathbf{x'})$, for $\mathbf{x'} \neq \mathbf{x}$.

A more numerically efficient recursive scheme is proposed by Le Gratiet and Garnier \cite{Le}. Suppose that the data sets have a nested structure, $i.e. D_1 \subseteq D_2 \subseteq \cdots \subseteq D_r$, this special scheme is derived by replacing the GP prior $Z_{t-1}(\mathbf{x})$ with the previous inference posterior $Z_{*t-1}(\mathbf{x})$. In this way, the problem becomes $r$ standard Gaussian process regression problems. So the resulting multi-fidelity posterior distribution can be denoted by $p(Z_t | \mathbf{y}_t, \mathbf{x}_t, Z_{*t-1}(\mathbf{x})), t = 1, \cdots, r$. The predictive mean and variance at each level are
\begin{equation}
\mu_{*t}(\mathbf{x}_*) = \rho \mu_{*t-1}(\mathbf{x}_*) + \mu_{\delta_t} + \mathbf{k}_{*n_t}\mathbf{K}^{-1}_t[\mathbf{y}_t - \rho \mu_{*t-1}(\mathbf{x}_t) - \mu_{\delta_t}]
\end{equation}
and 
\begin{equation}
\sigma_{*t}^2(\mathbf{x}_*) = \rho^2 \sigma_{*t-1}^2(\mathbf{x}_*) + \mathbf{k}_{**} - \mathbf{k}_{*n_t} \mathbf{K}^{-1}_t \mathbf{k}_{*n_t}^T
\end{equation}
where $n_t$ is the number of training points in data $D_t$ and $t$ denote the fidelity level.

\subsection{Multi-fidelity Gaussian process with nonlinear information fusion algorithm}
The above linear auto-regressive scheme is generalized in \cite{NARGP} as 
$$
Z_t(\mathbf{x}) = g_{t-1}(Z_{t-1}(\mathbf{x})) + \delta_t(\mathbf{x}),
$$
where $g_{t-1}$ is an unknown function quantifying the correlation between lower fidelity model and the higher one. Another GP prior is assigned to this function. However, the posterior distribution of $Z_t$ is not Gaussian anymore. This is the so-called deep GP in \cite{dGP2,dGP1}. At this point, the GP prior $Z_{t-1}$ is replaced by the previous inference result $Z_{* t-1}(\mathbf{x})$. In this way, using the additive structure of the scheme and the independence assumption between GPs $Z_{t-1}$ and $\delta_t$, which follows the construction assumption in \cite{autoregressive}, the above equation can be summarized as 
$$
Z_t(\mathbf{x}) = h_t(\mathbf{x}, Z_{*t-1}(\mathbf{x})),
$$
where $h_t \sim \mathcal{GP}(\mathbf{Z}_t|\mathbf{0}, k_t((\mathbf{x}, Z_{*t-1}(\mathbf{x})), (\mathbf{x}', Z_{*t-1}(\mathbf{x}'));\mathbf{\theta}_t))$. Essentially, this is a $(p+1)$ dimensional map which represents the relationship between the input space, the outputs of lower fidelity level model and the outputs of higher fidelity level model. 
The covariance kernel of the GP $h_t$ has a corresponding structure:
$$
k_{t_h} = k_{t_{rho}}(\mathbf{x}, \mathbf{x}';\theta_{t_{rho}}) \cdot 
k_{t_z}(Z_{*t-1}(\mathbf{x}), Z_{*t-1}(\mathbf{x}');\theta_{t_z}) + 
k_{t_{\delta}}(\mathbf{x}, \mathbf{x}';\theta_{t_{\delta}})
$$

The predictive posterior distribution of the first level of the above scheme is Gaussian but this is not the case for the remaining levels. So the predictive mean and variance are computed by using Monte Carlo integration of this following posterior distribution for $t \geq 2$:

\begin{equation}
\begin{split}
p(Z_{*t}(\mathbf{x}_*))  & := p(Z_t(\mathbf{x}_*, Z_{*t-1}(\mathbf{x}_*))|
Z_{*t-1}, \mathbf{x}_*,\mathbf{y}_t, \mathbf{x}_t)  \\
& = \int  p(Z_t(\mathbf{x}_*, Z_{*t-1}(\mathbf{x}_*))|\mathbf{x}_*,\mathbf{y}_t, \mathbf{x}_t) p(Z_{*t-1}(\mathbf{x}_*)) d\mathbf{x}_*
\end{split}
\end{equation}
More details can be found in \cite{NARGP}.

\subsection{Dimension reduction methods}
Dimension reduction is a popular topic in uncertainty quantification. Most dimension reduction methods are aimed at estimating the central sufficient dimension reduction subspace.

\begin{definition}{Dimension reduction:}
given a response scalar function $y = f(\bold{\xi})$, where $\xi=[\xi_1\dots\xi_p]^T$, a dimension reduction can be defined as a mapping from the $p$-dimensional input to a $d$-dimensional vector, i.e. $\eta = A\xi$, where $A\in R^{d \times p}, d<p$ and $AA^T = I$ is the identity matrix.
\end{definition}

\theoremstyle{definition}
\begin{definition}{Sufficient dimension reduction subspace (SDR subspace):}
let $X:\Omega\rightarrow R^p$ be a random vector. Let $Y:\Omega\rightarrow R$ be a random variable. The matrix $span(\beta)\in R^{p \times d}$ where $d<p$ is called a SDR subspace if 
$$ X\indep Y|\beta^TX$$
\end{definition}

\begin{definition}{Central SDR subspace:}
the central SDR subspace or the central subspace is defined as the intersection of all SDR subspaces, and is written as $S_{Y|X}$.
\end{definition}

Once an estimation of central subspace
matrix $A$ is obtained. We can define $\eta = A\xi$, then the function $y=f(\xi)$ can be rewritten into:
$$ y = f(\xi) \approx f(A^TA\xi) = f(A^T\eta) = g(\eta) $$ 
So the original model is reduced into a $d$-dimensional model where $d<p$ in this way. 

Sliced inverse regression(SIR) and Sliced average variance estimation(SAVE) are two commonly used methods to estimate central subspace by approximating the conditional expectation $E(\xi|Y)$ and conditional variance $E(\xi\xi^T|Y)$. The detail information and the software package implementation is available at \cite{SReduR}. The two methods are shown in Algorithm 1 and Algorithm 2.
In all numerical examples, we use SAVE method to conduct necessary computations but it can be easily replaced with SIR or other similar methods. 

\begin{algorithm}
\caption{Sliced Inverse Regression(SIR)}
\begin{algorithmic}
\STATE 1. Compute the sample mean and sample variance:$$\hat{\mu}=E_n(X),\hat{\sigma}=var_n(X).$$ and compute the standardized random vectors $$Z_i=\hat{\Sigma}^{-1/2}(X_i-\hat{\mu}), i=1,\dots,n.$$
\STATE 2. Discretize $Y$ as $\hat{Y} = \sum_{h=1}^H hI(Y\in J_h)$, where a collection of intervals $\{ J_1,\dots,J_h\}$ is a partition of $Y_i$.
\STATE 3. Approximate $E[Z|\hat{Y}\in J_h]$ or $E[Z|Y\in J_h]$ by $$ E_n(Z|Y\in J_h) = \frac{E_n[ZI(Y\in J_h)]}{E_n[I(Y\in J_h)]}, l=1,\dots,H $$
\STATE 4. Approximate $var[E(Z|\hat{Y})]$ by $$ M=\sum_{h=1}^H E[I(Y\in J_h)]E_n(Z|Y\in J_h)E_n(Z^T|Y\in J_h)$$
\STATE 5. Let $\hat{v}_1,\dots,\hat{v}_d$ be the first $d$ eigenvectors of $M$, let $\hat{\beta}_k = \hat{\Sigma}^{-1/2}\hat{v}_k$,$k=1,\dots,d$. The SDR predictors are $[\hat{\beta}_1^T(X_1-\hat{\mu}),\dots, \hat{\beta}_d^T(X_d-\hat{\mu})]$.
\end{algorithmic}
\end{algorithm}

\begin{algorithm}
\caption{Sliced Average Variance Estimation(SAVE)}
\begin{algorithmic}
\STATE 1. Standardize $X_1,\dots, X_n$ to obtain $Z_i$ as in Algorithm 1.
\STATE 2. Discretize $Y$ as $\hat{Y} = \sum_{h=1}^H hI(Y\in J_h)$, where a collection of intervals $\{ J_1,\dots,J_h\}$ is a partition of $Y_i$.
\STATE 3. For each slice $J_h$, compute the sample conditional variance of $Z$ given $Y\in J_h$: $$ var_n(Z|\hat{Y}=h)= \frac{E_n[ZZ^TI(\hat{Y}=h)]}{E_n[I(\hat{Y}=h)]} $$
\STATE 4. Compute the sample version of $M$:$$ M = H^{-1}\sum_{h=1}^H E_nI(\hat{Y}=h)[I_p - var_n(Z|\hat{Y}=h)]^2 $$
\STATE 5. Let $\hat{v}_1,\dots,\hat{v}_d$ be the first $d$ eigenvectors of $M$, let $\hat{\beta}_k = \hat{\Sigma}^{-1/2}\hat{v}_k$,$k=1,\dots,d$. The SDR predictors are $[\hat{\beta}_1^T(X_1-\hat{\mu}),\dots, \hat{\beta}_d^T(X_d-\hat{\mu})]$, where $\hat{\mu} = E_n(X)$.
\end{algorithmic}
\end{algorithm}

A challenge in the dimension reduction problem is the determination of the reduced dimension $d$. In this paper we choose the Bayesian information criterion (BIC) introduced in \cite{SReduR}. Let $\lambda_1\geq\lambda_2\geq\dots\geq\lambda_p$ be the eigenvalues of the $\hat{V} + I $, where $\hat{V}$ is the co-variance matrix in SIR/SAVE algorithm. Assume $\hat{V}$ is positive semi-definite, we have $\lambda_i\geq 1$ for all $i\leq p$. Let 
\begin{equation}
    G(k) = \frac{n}{2} \sum_{l=1+k}^p (log\lambda_l + 1 -\lambda_l) - C_nk(2p-k+1)/2 
\end{equation}
where $C_n$ is a sequence satisfying the condition in Theorem 2 of \cite{Ona}. Then the number of dimensions $d$ is approximated by:
$$ d = arg max{G(k): k = 1,\cdots, p-1} $$

In the last step of our proposed model, if the parameter $flag = 1$, the number of original dimensions $p$ is reduced to some number slightly larger than $d$ depending on the number of high-fidelity training data. Then the Gaussian process dimension reduction technique is applied to compute another reduction matrix in order to reduce the number of dimensions to $d$.

\subsection{Gaussian process dimension reduction technique}
A similar approach is proposed in \cite{GPbDR} and based on a novel covariance function of Gaussian process,
$$
k_s(\bold{x}, \bold{x}'; \bold{W},\bold{\phi}) = k_d(\bold{W}^T \bold{x}, \bold{W}^T \bold{x}';\bold{\phi})
$$
where $k_s$ is a standard covariance function and $k_d$ is the corresponding covariance function on a low-dimensional space. So, the inputs are first projected to a low-dimensional space before feeding to the Gaussian process covariance function. Note that the newly constructed kernel $k_s$ has both the projection matrix $\bold{W}$ and the original kernel parameters $\bold{\phi}$ as its parameters. Those hyper-parameters are joint optimized in Gaussian process regression using maximizing the marginal log-likelihood function. The process for this method is shown in Algorithm 3.

\begin{algorithm}
\caption{Gaussian process dimension reduction technique}
\begin{algorithmic}
\STATE 1. Input: high-fidelity data set $\{X_H,y_H\}$ and validation data set $\{X_T,y_T\}$, iteration number $N$, input dimension $s$, output dimension $d$, initial guess of reduction matrix $A_0$ and hyper-parameters $\bold{\theta}_0$ for GP kernel.
\STATE 2. Project the high-fidelity data set $\{X_H,y_H\}$ to a low-dimensional space using $A_0$ to get a new data set $\{\hat{X_H},y_H\}$.
\STATE 3. Build a Gaussian process model with $\{\hat{X_H},y_H\}$ from previous step and optimization the model using maximum likelihood method with the initial guess $A_0$ and $\bold{\theta}_0$.
\STATE 4. Fix the reduction matrix $A$ as parameters of the GP kernel and optimize the hyper-parameters $\bold{\theta}$. 
\STATE 5. Unfix parameters $A$ and fix the hyper-parameters $\bold{\theta}$, then optimize $A$ again.
\STATE 6. Repeat step 4 and 5 for $N$ times.
\STATE 7. Output: a reduction matrix $M_2$ with size $s \times d$ which is the optimal parameter $A$ in the GP optimization process.
\end{algorithmic}
\end{algorithm}

For small set of high-fidelity training data, the method in \cite{GPbDR} can not be directly applied to the original high-dimensional problem to obtain accurate results. For instance, if the number of original dimensions is $p$ and true number of reduced dimensions is $d$. Then we have $p \times d$ additional parameters to optimize except original hyper-parameters in a standard Gaussian process regression. In our method, the number of original dimensions $p$ is first reduced to $s$ with our rotated multi-fidelity Gaussian process model. Then Gaussian process dimension reduction technique can be effectively applied to reduce the number of dimensions from $s$ to $d$. Here, $s$ depends on the number of high-fidelity samples in the problem and can be chosen by the user. In our numerical examples, the number of dimensions $s$ is chosen to be $3$ to better demonstrate our method. For other values of $s$, it can be implemented similarly.

\subsection{Bayesian active learning}
Active learning aims at maximizing information acquisition with limited data. It is also known as optimal experimental design or sequential design in statistic literatures. As Sverchkow and Craven(2017) stated in \cite{ReAC}, informative experiments are first proposed according to the hypotheses generated from the model. Then the model is updated by the data obtained from the experiments. In this way, the model is gradually improved from such an iterative process which is called active learning. More recent works on this topic includes \cite{AL1}, \cite{AL3}, \cite{AL2}.

In the problem setting of this paper, the training data come from the low-fidelity models are rich and easy to obtain and it is denoted by,

$$D_L  = (X_L, Y_L) $$

But data from the high-fidelity model are time consuming or very expensive, which is denoted by,

$$D_H  = (X_H, Y_H)$$
In order to make accurate inference, more training data points need to be selected and added to the high-fidelity data set. The candidate pool is chosen to be the low-fidelity observation set. With the help of Bayesian active learning, the additional data points to augment the original high-fidelity observations can be efficiently determined. This can greatly reduce the model uncertainties in the problem setting under limited budget.

Assuming the training data set consists of $N_L$ low-fidelity observations $D_L$ and $N_H$ high-fidelity observations $D_H$. So the training data set $D$ can be expressed as,
$$
D = \{ D_L, D_H\}
$$
This represents the current state of knowledge and a multi-fidelity model can be built as stated in the previous sections. Now, the most informative sample in the low-fidelity observations is picked by maximizing an acquisition function $a_N(x)$,
\begin{equation}
    \mathbf{x}_{N_H+1} = \textit{argmax}_{\mathbf{x} \in D_L} a_N(\mathbf{x})
\end{equation}

The acquisition function actually quantifies how much information we can get to evaluate or perform an expensive experiment at this data point. Then $(x_{N_H+1}, y_{N_H+1})$ is added to the high-fidelity observation set $D_H$. At this point, the process stops if a pre-set problem related stop criterion is achieved. Otherwise, the process repeats iteratively until it satisfies the stop criterion or reaches the maximal number of times permitted. 

There are several common acquisition functions in Bayesian active learning, including maximum upper interval, probability of improvement and expected improvement, based on different problem settings. In our multi-fidelity setting, the prediction variance quantifies how much uncertainties the model has for the current predictions. This guides us to choose the acquisition function to be the predictive variance of the model:
\begin{equation}
a_N(\mathbf{x}) = \sigma^2_*(\mathbf{x})
\end{equation}

As for the stop criterion, the Bayesian active learning process stops if the relative error of the predictions of the proposed rotated multi-fidelity model on the test set is less than a chosen small value $\eta$.


\section{Algorithm}
In this section, the rotated multi-fidelity Gaussian process(RMFGP) model and dimension reduction process built on it is introduced. In the proposed method and the numerical results in the following, we consider two layers of fidelity. The deeper layer problems share the similar principle. Given the low and high-fidelity training data $\{X_L,y_L\}$, $\{X_H,y_H\}$ and the test data $\{X_T, y_T\}$, Algorithm 4 summarizes the process.

\begin{algorithm}
\caption{Rotated multi-fidelity Gaussian process model(RMFGP)}
\begin{algorithmic}
\STATE 1. Input: low-fidelity data sets $\{X_L,y_L\}$,  high-fidelity set $\{X_H,y_H\}$ and validation data set $\{X_T,y_T\}$, threshold $\xi$, maximum iteration number $I$, reduction parameter $flag = 0$ or $1$.  
\STATE 2. Apply SAVE method(Alg. 2) only on the low-fidelity data to compute the first rotation matrix $A_T$ to extract the principle direction information in the low-fidelity data. Then apply $A_T$ to all $X= (X_L, X_H, X_T)$.
\STATE 3. Perform NARGP on new training data $\{\hat{X_L},y_L\}$, $\{\hat{X_H},y_H\}$ from step 1 to get the prediction $\hat{y_T}$ at $\hat{X_T}$. Then perform SAVE method again on new $\{\hat{X_T}, \hat{y_T}\}$ to compute the rotation matrix $\hat{A}$ and apply $\hat{A}$ to all $X$s.
\STATE 4. Check whether the threshold of the generalization error meet. If not, perform Bayesian active learning method to locate $x^*$ where the prediction variance achieves maximum. Sample $\{x^*, y_H^*\}$ and add them into the high-fidelity training set. 
\STATE 5. Repeat step 2 and 3 until the generalization error threshold fulfilled or the maximum iteration number is reached.
\STATE 6. Compute the rotation matrix $M_1 = A_T\prod_i \hat{A_i}$. and build the rotated model.
\STATE 7. If dimension reduction parameter $flag = 1$, compute the intrinsic dimension $d$ through BIC and compute the reduction matrix $\hat{M}_1$ consisting of the first $s$ principle columns of $M_1$, where $d < s < p$.
\STATE 8. Apply Gaussian processes dimension reduction technique(Alg. 3) to compute reduction matrix $M_2$ which is $s \times d$. Then build the reduced model with final reduction matrix $M = \hat{M}_1 M_2$.
\end{algorithmic}
\end{algorithm}







\section{Numerical Results}
In this section, four examples are present to demonstrate the effectiveness of RMFGP model. We first measure the accuracy of the approximated rotation matrix $\hat{A}$ by the distance metric defined in \cite{Ona}:
\begin{equation}
m(A, \hat{A}) = ||P - \hat{P}|| 
\end{equation}
where $A$ and $\hat{A}$ are the true and estimated central subspace matrices, $P$ and $\hat{P}$ are projection matrices of $A$ and $\hat{A}$, $|| \cdot ||$ is the Frobenius norm. To illustrate the effectiveness of our method, the reduction matrix computed by RMFGP when $flag=1$ and the method introduced in \cite{BIRsd} are compared using the same number of high-fidelity samples.

Once the rotation matrix is computed, new training and test set are generated to build and evaluated a new Gaussian process model. To compare the accuracy between final surrogate models under the fair setting, two cases are considered depending on the reduction parameter $flag$. If $flag = 0$, the proposed method and the standard Gaussian process regression(GPR) are performed on the origin and rotated data set respectively and the results are compared. If $flag = 1$, the reduced model is performed on the data set with the reduction matrix found in RMFGP model. It is then compared to the GP-SAVE method introduced in \cite{BIRsd}. The accuracy of the models is measured by the relative error defined in \cite{EnSp}:
\begin{equation}
    e = \frac{||u - \hat{u}||_2}{||u||_2}
\end{equation}
where $u$, $\hat{u}$ are the exact and approximated values of the high-fidelity model on the test set and $|| \cdot ||_2$ is the $L_2$ norm. 

\subsection{Linear example: Poisson's equation}
The first example illustrates the situation where the high-fidelity function has a linear relationship with the low-fidelity function. This can be seen as the simplest form of relationship  between the high-fidelity and low-fidelity functions. Consider the equations:
$$ f_H(\bold{x}) = sin(\pi (x_1 + x_3)) + sin(\pi (x_1 + x_2)) + 2 $$
$$ f_L(\bold{x}) = f_H(\bold{x}) + x_3 x_4 x_5 x_6 $$
where $x_i$, $i = 1,\dots,6$, are i.i.d uniformly distributed random variables on $[0,1]$. In this example, the high-fidelity and low-fidelity function are a solution of a Poisson's equation according to different force terms respectively,
\begin{equation}
    \nabla^2 f = h(x)
\end{equation} 
$$ h_H(x) = 2\pi^2sin(\pi(x_1 + x_3)) + 2\pi^2sin(\pi(x_1 + x_2)) $$
$$ h_L(x) = h_H(x) + \frac{1}{6}(x_3^3x_4x_5x_6 + x_3x_4^3x_5x_6 + x_3x_4x_5^3x_6 + x_3x_4x_5x_6^3)$$
From this point of view, one can also think the difference between high and low-fidelity function comes from the complicated high dimensional noise in the force term $h_L$.
The input dimension is $p=6$. It is easily seen from the expression that the inputs for $f_H$ are actually in a subspace of dimension $d = 2$. The actual dimension reduction matrix $A = span\{\beta_1, \beta_2\}$, where $\beta_1 = (1,0,1,0,0,0)^T, \beta_2 = (1,1,0,0,0,0)$. The number of low-fidelity training points is set to be $N_L = 200$ and the number of test points is set to be $N_T = 500$. For all cases, the high-fidelity sample size starts at $(N_H-10)$ and two iterations in Bayesian active learning process are involved with $5$ samples added per iteration. The results are shown in Table $1$ and Table $2$.

\begin{table}[h!]
\centering
\begin{tabular}{||c c c c c||} 
 \hline
              &  $N_H=25$      &  $N_H=30$      &    $N_H=35$    & $N_H=40$ \\ [0.5ex] 
  \hline
 RMFGP ($flag=1$)  & 0.133262      &    0.112705   &    0.066355  &  0.043337    \\ 
 \hline
 GP-SAVE    &  0.396780      &    0.243222    &   0.221707    &   0.202306    \\  
 \hline
\end{tabular}
\caption{Accuracy of the dimension reduction matrix measured by the metric $m(A,\hat{A})$ with different number of high-fidelity samples for linear examples - Equation (11). The number of low-fidelity samples is fixed to be $N_L = 200$. There are two iterations to add high-fidelity samples in Bayesian active learning process with $5$ points added per iteration. For RMFGP method, the number of dimensions of the inputs is first reduced to $s = 3$ from the number of original dimensions $p=6$. Then Gaussian process dimension reduction technique is applied to reduce the dimension from $s=3$ to $d=2$. The additional number of hyper-parameters needed to optimized in this step is $6$.}
\label{table:1}
\end{table}

\begin{table}[h!]
\centering
\begin{tabular}{||c c c c c||} 
 \hline
                     & $N_H= 25$   &   $N_H=30$     &    $N_H=35$    &    $N_H=40$  \\ [0.5ex] 
 \hline
 RMFGP ($flag=0$) & 0.051382   &   0.030260    &    0.019012    &    0.007531 \\ 
 \hline
 GP     & 0.060513  &  0.043898    &    0.027694    &   0.023358  \\
 \hline
 RMFGP ($flag=1$) & 0.051358  &   0.039532   &    0.026999   &  0.020309 \\ 
 \hline
 GP-SAVE   & 0.084310   &  0.077906     &   0.072945     &  0.066365   \\  
\hline
\end{tabular}
\caption{Relative error $e$ of RMFGP model compared to standard GP for linear examples - Equation (11). If $flag=0$, the inputs are simply rotated by the rotation matrix from RMFGP model before fed into a new GP surrogate model. It is compared to a standard GP model. If $flag=1$, the inputs are reduced to dimension $d=2$. For comparison, the inputs for the standard GP are reduced to dimension $d=2$ by a reduction matrix computed by SAVE method using the same number of high-fidelity training points.}
\label{table:2}
\end{table}

Table $1$ summarizes the distance between the approximate reduction matrix and true reduction matrix measured by (9) based on four different sample sizes with the reduction parameter $flag = 1$. The accuracy of estimating the central subspace for both methods increases with the sample size showing that the methods are consistent. The proposed method performs better across all sample sizes. This is because of the information provided by the low-fidelity data and the improvement of the prediction performance through the active learning. This indicates a better accuracy of our method than traditional dimension reduction method, especially when we have only limited budget for acquiring high-fidelity data. Table $3$ is the BIC results illustrating the value of $G(k)$ in Equation (6) with different k. The approximated reduced dimension is $\hat{d} = 2$ according to (6), which is the same as the one obtained from the expression of $f_H$ directly. 

\begin{table}[h!]
\centering
\begin{tabular}{||c c c c c c c||} 
 \hline
              &  k=1         &  k=2                    &   k=3                   & k=4        & k=5      & k=6       \\ [0.5ex] 
 \hline
  G(k)        &  0.9366920 &   $\textbf{0.9592096}$    &    0.9272198  &  0.8946359   &  0.8609161   & 0.8267321     \\  
 \hline
\end{tabular}
\caption{BIC: G(k) for linear examples - Equation (11)}
\label{table:3}
\end{table}

Table $2$ shows the relative errors on the test set as measured by (10) for each combination of the four models involved and four different sample sizes $n_H=25,30,35,40$. The first two rows represent the comparison between the proposed method and the standard Gaussian process regression(GPR) when the dimension reduction parameter $flag = 0$. The last two rows represent the comparison between the proposed method when $flag = 1 $ and the GP method with the dimension reduction using SAVE method (GP-SAVE). For our RMFGP method, we first project the original inputs $X$ from the number of dimensions $p=6$ to the reduced dimension $s=3$. Then, Gaussian process dimension reduction technique is utilized to further reduce the number of dimensions from $s=3$ to $d=2$. Notice that there are $6$ additional parameters we need to optimize in this step, so the requirement for number of high-fidelity data can be relaxed in order to achieve certain accuracy. All models illustrate the consistency that the relative error decreases with the increase value of $N_H$. The proposed RMFGP method performs better in both situations. Figure $2$ shows the mean square error(MSE) of different models with respect to the size of high-fidelity data. 
The red curve represents the proposed methods and the blue curve represents the comparison methods. Both figures reveal the fact that the proposed method has a smaller MSE and a faster convergence, especially when $N_H$ is small.

\begin{figure}[h]
\centering
\subfloat[a][]{
\includegraphics[width=0.45\textwidth]{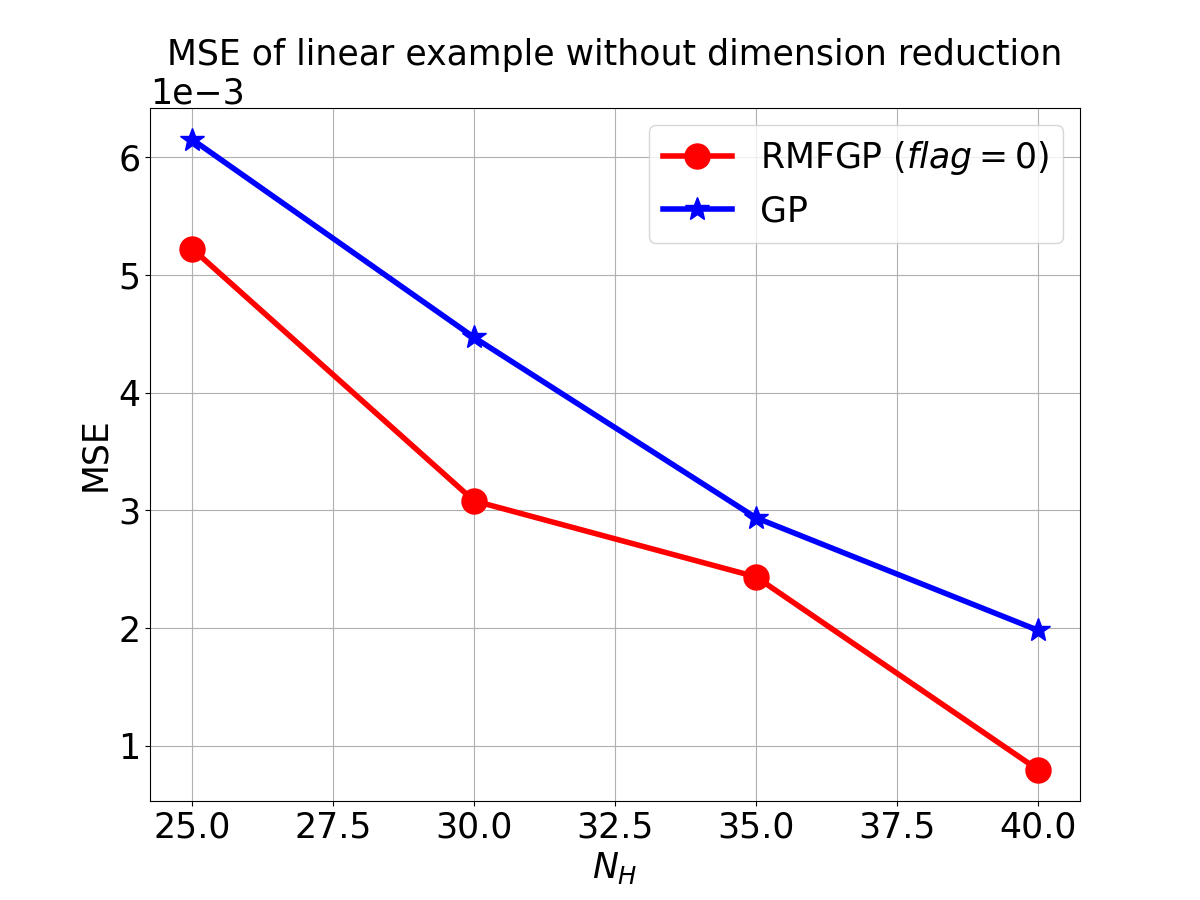}
\label{fig:MSE_linear1}}
\qquad 
\subfloat[b][]{
\includegraphics[width=0.45\textwidth]{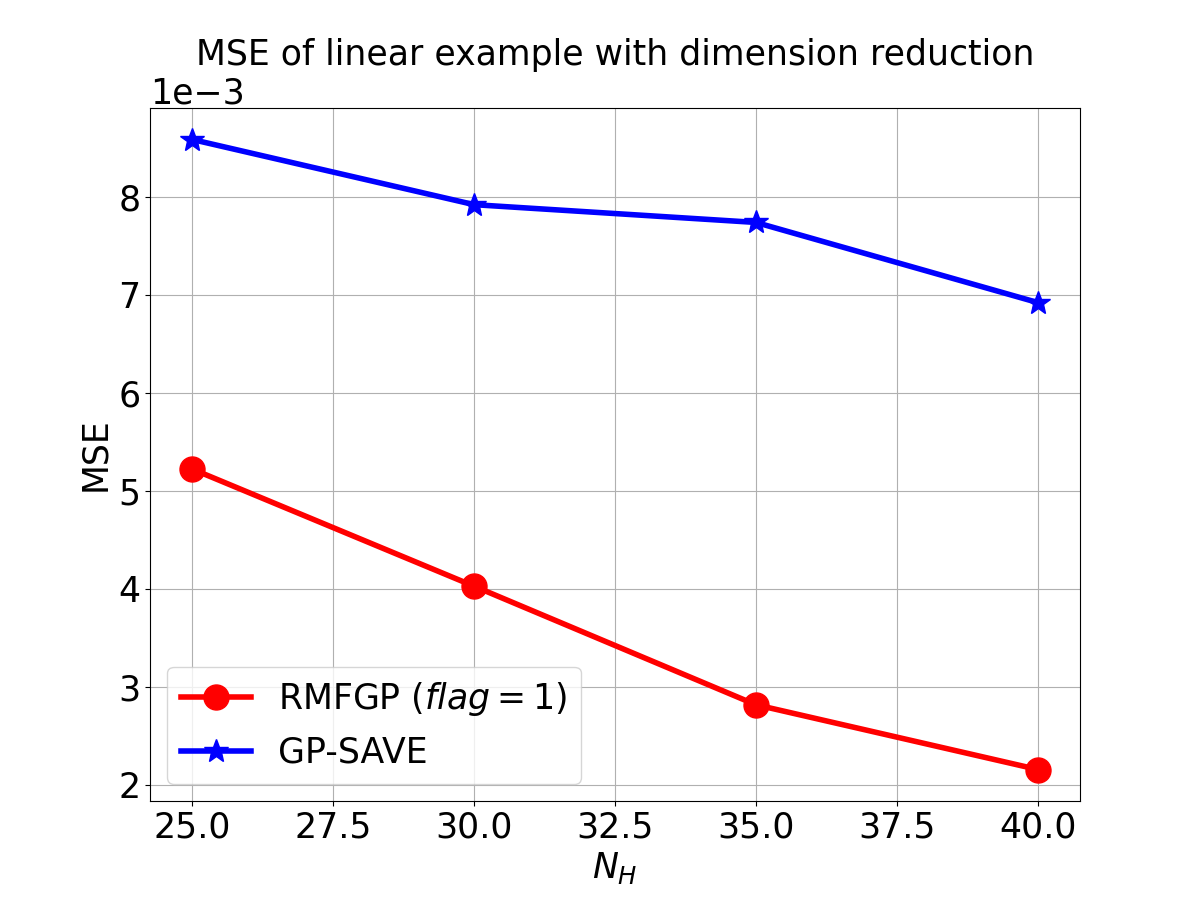}
\label{fig:MSE_linear2}}
\caption{MSE of linear examples - Equation (11): (a) Models without dimension reduction: RMFGP ($flag=0$) vs. GP; (b) Models with dimension reduction: RMFGP ($flag=1$) vs. GP-SAVE. For both (a) and (b), axis x is the number of high-fidelity samples $N_H$ and the number of low-fidelity sample $N_L$ is fixed to be 200. The number of original dimensions is $p = 6$. }
\label{fig:MSE_linear}
\end{figure}

Figure $3$ represents the correlation between the prediction and the true observation at $N_H = 30$. The figure on the left demonstrates the correlation of the RMFGP model when $flag = 0$ and GP with the original data $X$, while the figure on the right shows the correlation of the RMFGP model when $flag = 1$ and GP-SAVE with the reduced input $\hat{X}$. The red dots represent the results of RMFGP. The blue squares are the results of the comparison methods. The black solid line represents the perfect correlation between the predictions and true observations. As shown in Figure $3$, the red dots stay close to the perfect correlation while the blue squares are around but somehow off the black solid line, which indicates a better prediction on the test set for the proposed method. Hence, we can conclude that the proposed method estimate both the central subspace and the model predictions better than the traditional methods. The user can decide whether to use the rotated model or the reduced model based on the needs of the real application by controlling the reduction parameter $flag$ in the inputs of Algorithm 4.

\begin{figure}[h]
\centering
\subfloat[a][]{
\includegraphics[width=0.45\textwidth]{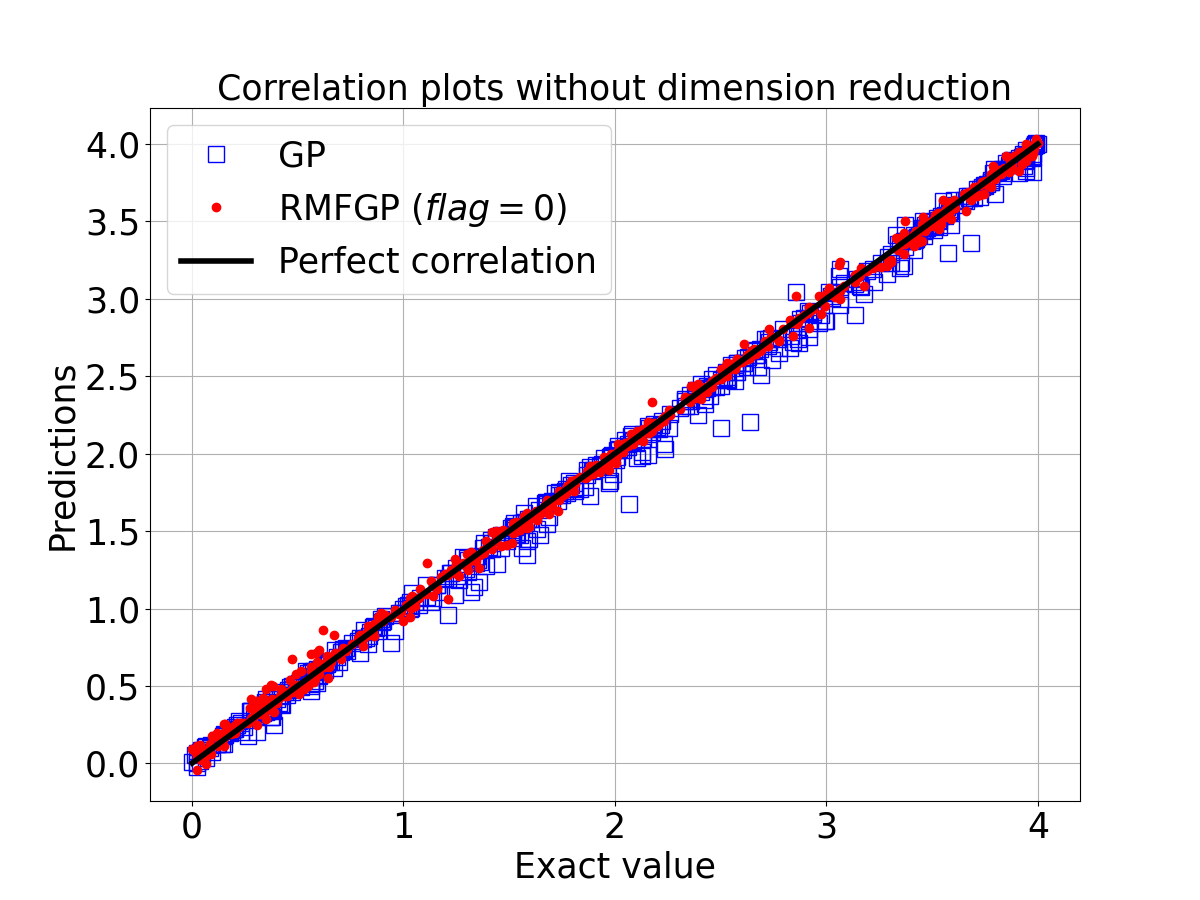}
\label{fig:LinearCorSAVE1-30}}
\qquad
\subfloat[b][]{
\includegraphics[width=0.45\textwidth]{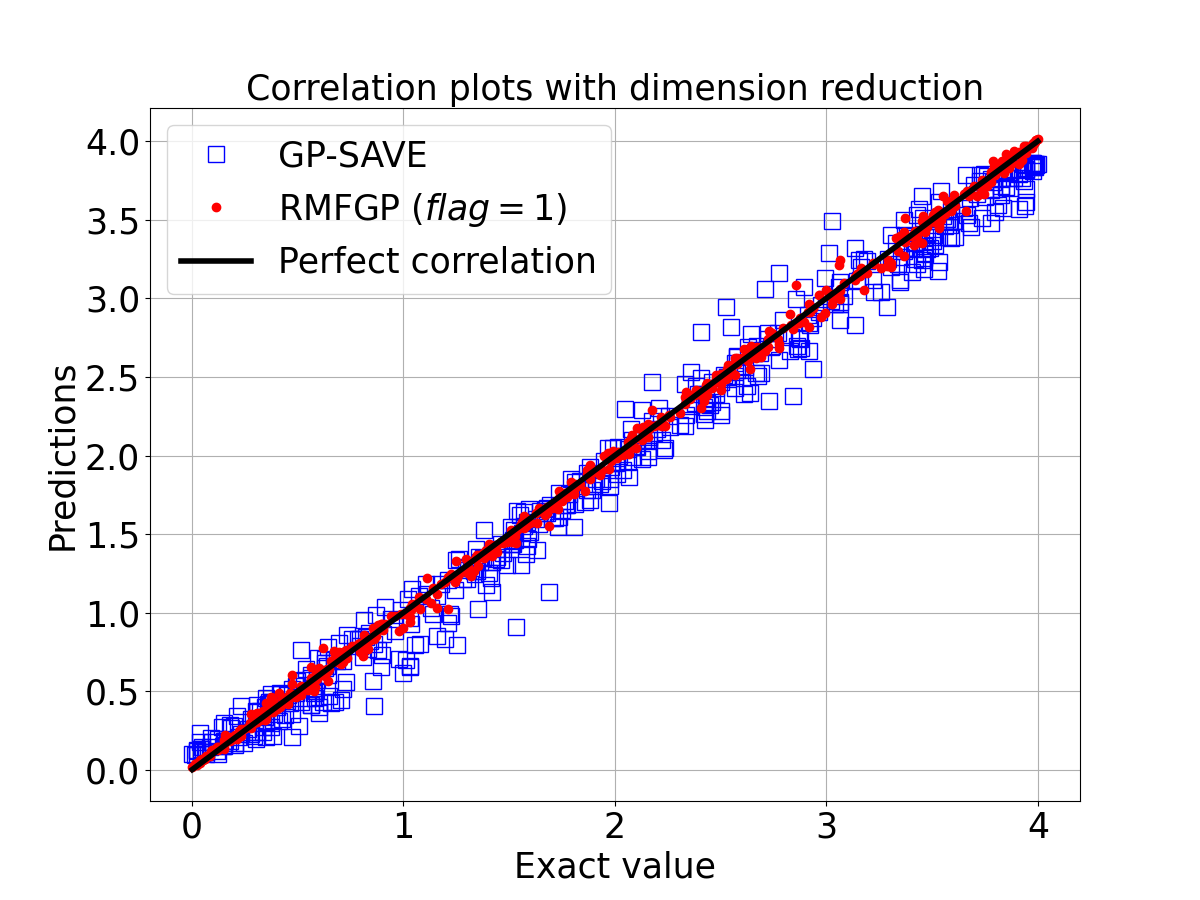}
\label{fig:LinearCorSAVE2-30}}

\caption{Correlation plots of linear examples - Equation (11) at $N_H=30$: (a) RMFGP ($flag=0$) vs. GP; (b) RMFGP ($flag=1$) vs. GP-SAVE. For all figures, axis x is the exact values and axis y is the predictions at the test points. The black solid line is the perfect correlation. }
\label{fig:LinearCor}
\end{figure}

\subsection{Nonlinear example}
The second example demonstrates that RMFGP can predict model with more complicated relationship between high and low-fidelity data, which can not be predicted by the linear auto-regressive model. Consider the following function:
$$ f_H(\bold{x}) = exp(0.2 \sum_{i=1}^{10} x_i) $$
$$ f_L(\bold{x}) = x_4 f_H(\bold{x}) $$
where $x_i$, $i = 1,\dots,10$ are i.i.d uniformly distributed random variables on $[0,1]$. Based on the expression of $f_H$ and $f_L$, the actual number of dimensions of the inputs for low-fidelity function is $2$ while it is $1$ for high-fidelity function. This can happen in the real world applications where the low-fidelity data contain various noises. The original number of input dimensions is $p=10$. The actual dimension reduction matrix $A = span\{\beta_1\}$, where $\beta_1 = (1,1,1,1,1,1,1,1,1,1)^T$. In this experiment, the number of low-fidelity training points used is $N_L = 200$ and the number of test points is set to be $N_T = 500$. For all the cases, the start number of high-fidelity samples is $(N_H - 5)$, the active learning scheme is then employed to add $2$ points in first iteration and $3$ points in last iteration before it reaches the stopping criterion.

Table $4$ shows the distance as measured by (9) for this experiment. As expected, both RMFGP and GP illustrate the consistency. Note that RMFGP gains a high accuracy of the estimated central subspace with a relatively small high-fidelity data set. 
In this example, the low-fidelity data contains full information about high-fidelity data but with some noises as a multiplier in front of it. The $G(k)$ values are presented in Table $5$. It reaches maximum at $k=1$, which indicates the estimated reduced dimension is $\hat{d}=1$. It is the same as the one obtained from the expression of $f_H$ directly. It suggests the capability of RMFGP to identify the intrinsic dimension under the effect of some noises. So if the reduction parameter $flag=1$, we first reduced the number of dimensions from $p=10$ to $s=3$ and then apply the Gaussian process dimension reduction technique to further reduce the dimension to $d=1$. 

\begin{table}[h!]
\centering
\begin{tabular}{||c c c c c||} 
 \hline
              &  $N_H=10$      &  $N_H=15$      &    $N_H=20$   & $N_H=25$ \\ [0.5ex] 
  \hline
 RMFGP ($flag=1$) &  0.375618     &   0.217788  &   0.032944    &  0.019125   \\
 \hline
 GP-SAVE    & 1.234884    &   1.004467   &  0.177209  &   0.080237   \\  
 \hline
\end{tabular}
\caption{Accuracy of the dimension reduction matrix measured by the metric $m(A,\hat{A})$ with different number of high-fidelity samples for nonlinear examples. The number of low-fidelity samples is fixed to be $N_L = 200$. There are two iterations to add high-fidelity samples in Bayesian active learning process with $2$ points added in first iteration and $3$ points added in the second. For RMFGP method, the number of dimensions of the inputs is first reduced to $s = 3$ from the number of original dimensions $p=10$. Then Gaussian process dimension reduction technique is applied to reduce the dimension from $s=3$ to $d=1$. The additional number of hyper-parameters needed to optimized in this step is $3$.
}
\label{table:4}
\end{table}

\begin{table}[h!]
\centering
\begin{tabular}{||c c c c c c||} 
 \hline
              &  k=1         &  k=2                    &   k=3                   & k=4        & k=5 \\ [0.5ex] 
 \hline
  G(k)        &  $\textbf{0.9622132}$ &   0.9355113    &    0.9061465  &  0.8764734   &  0.8466059   \\
 \hline
  & k=6    & k=7 & k =8 &
              k=9 & k = 10\\ [0.5ex]
\hline
    G(k) &    0.8163634 & 0.7852849 & 0.7539464 &
  0.7221620 & 0.6893825 \\  
 \hline       
\end{tabular}
\caption{BIC: G(k) for nonlinear examples}
\label{table:5}
\end{table}

Table $6$ and Figure $4$ are the relative error and the MSE plots of four models with different sizes of $N_H$. RMFGP outperforms GP on both cases which $flag=0$ or $1$. It has smaller errors on all sample sizes and it converges faster than the comparison method. This concludes the RMFGP dimension reduction model can successfully identify the accurate central subspace and achieve a low relative error.

\begin{table}[h!]
\centering
\begin{tabular}{||c c c c c||} 
 \hline
                     & $N_H= 10$   &   $N_H=15$     &    $N_H=20$     &    $N_H=25$  \\ [0.5ex] 
 \hline
 RMFGP ($flag=0$) & 0.037386   &   0.008324    &    0.001096    &    0.000810 \\ 
 \hline
 GP                  & 0.171120   &   0.139623     &   0.034945    &   0.024370 \\
 \hline
  RMFGP ($flag=1$) & 0.080942   &   0.045254    &    0.006634   &   0.003374 \\ 
 \hline
 GP-SAVE            & 0.175399  &   0.133578    &   0.026675    &   0.021621 \\  
\hline
\end{tabular}
\caption{Relative error $e$ of RMFGP model compared to standard GP for nonlinear examples. If $flag=0$, the inputs are simply rotated by the rotation matrix from RMFGP model before fed into a new GP surrogate model. It is compared to a standard GP model. If $flag=1$, the inputs are reduced to dimension $d=1$. For comparison, the inputs for the standard GP are reduced to dimension $d=1$ by a reduction matrix computed by SAVE method using the same number of high-fidelity training points.}
\label{table:6}
\end{table}

\begin{figure}[h]
\centering
\subfloat[a][]{
\includegraphics[width=0.45\textwidth]{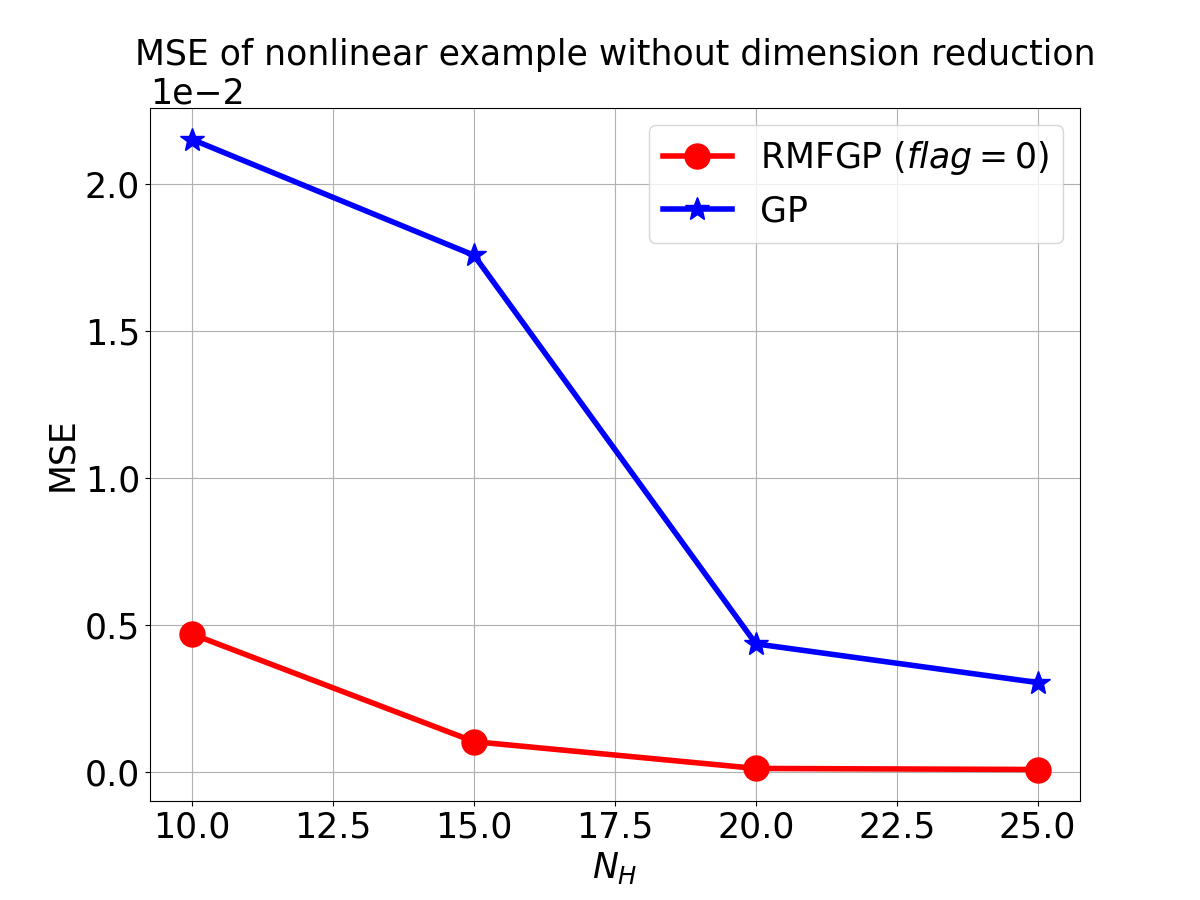}
\label{fig:mse_nonlinear1}}
\qquad
\subfloat[b][]{
\includegraphics[width=0.45\textwidth]{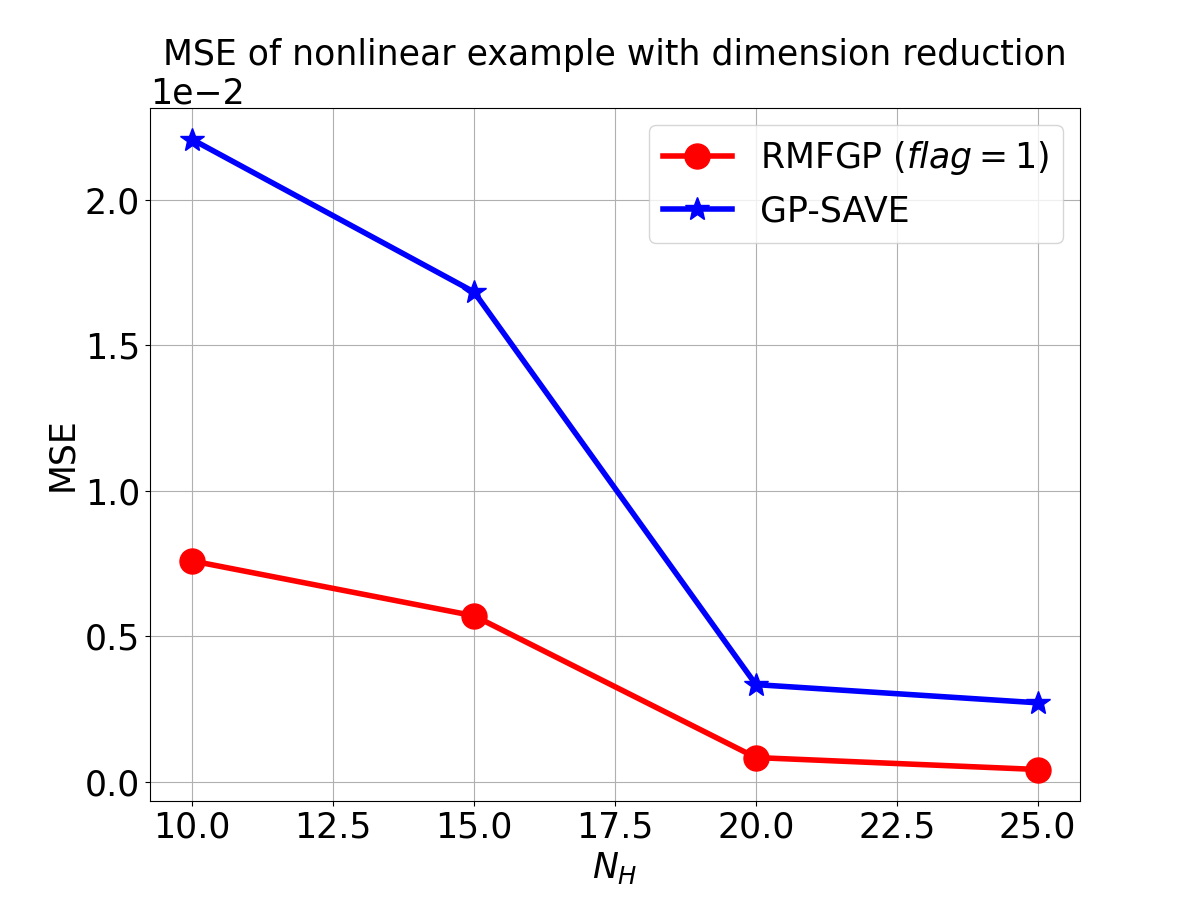}
\label{fig:mse_nonlinear2}}
\caption{MSE of nonlinear examples: (a) Models without dimension reduction: RMFGP ($flag=0$) vs. GP; (b) Models with dimension reduction: RMFGP ($flag=1$) vs. GP-SAVE. For both (a) and (b), axis x is the number of high-fidelity samples $N_H$ and the number of low-fidelity sample $N_L$ is fixed to be 200. The number of original dimensions is $p = 10$. }
\label{fig:mse_nonlinear}
\end{figure}

The correlation between the predictions and the true observations is presented in Figure $5$. The number of high-fidelity training data used is $N_H = 20$. The color settings are the same as that in Example $4.1$. The proposed method acts well under all situations since the red dots are very close to the perfect correlation line while GP can not predict the test set well with small $N_H$.

\begin{figure}[h]
\centering
\subfloat[a][]{
\includegraphics[width=0.45\textwidth]{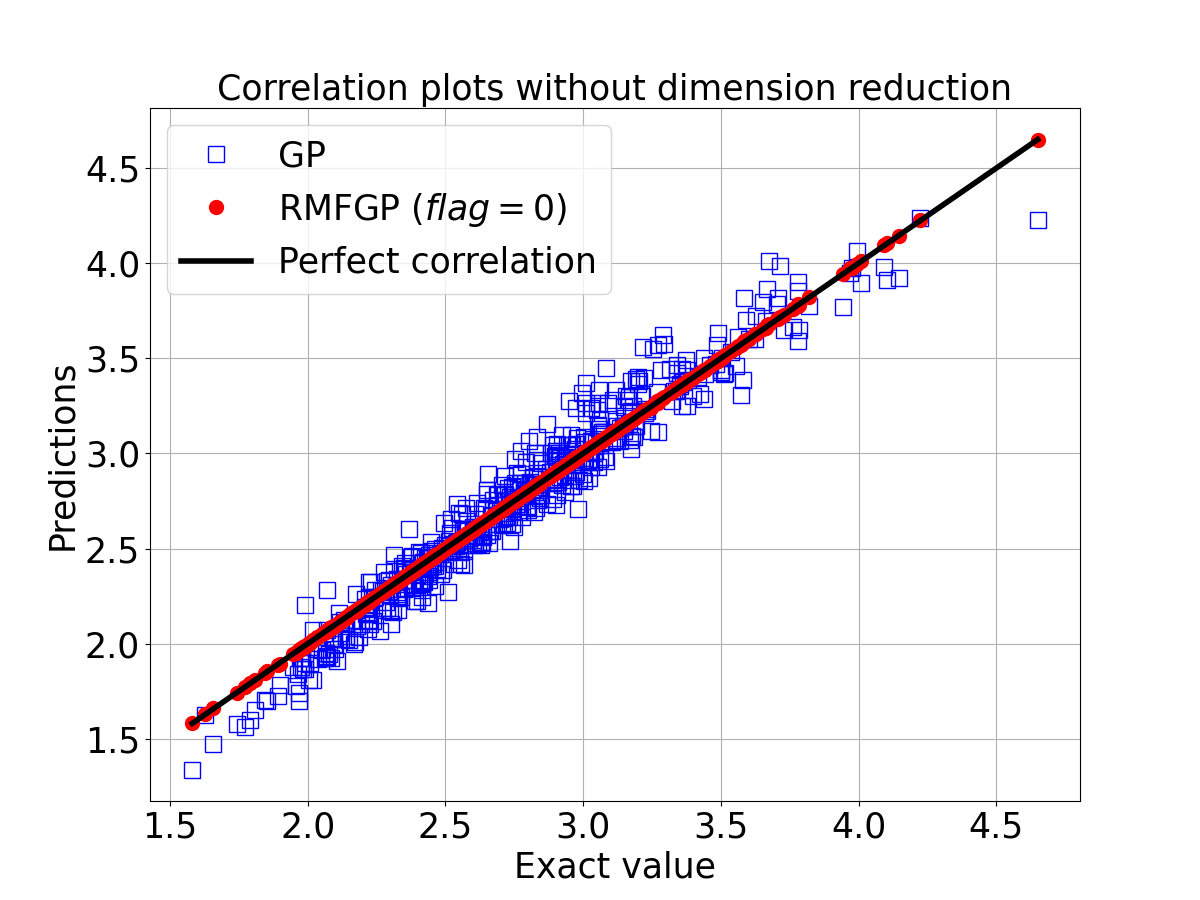}
\label{fig:NonlinearCorSAVE1-20}}
\qquad
\subfloat[b][]{
\includegraphics[width=0.45\textwidth]{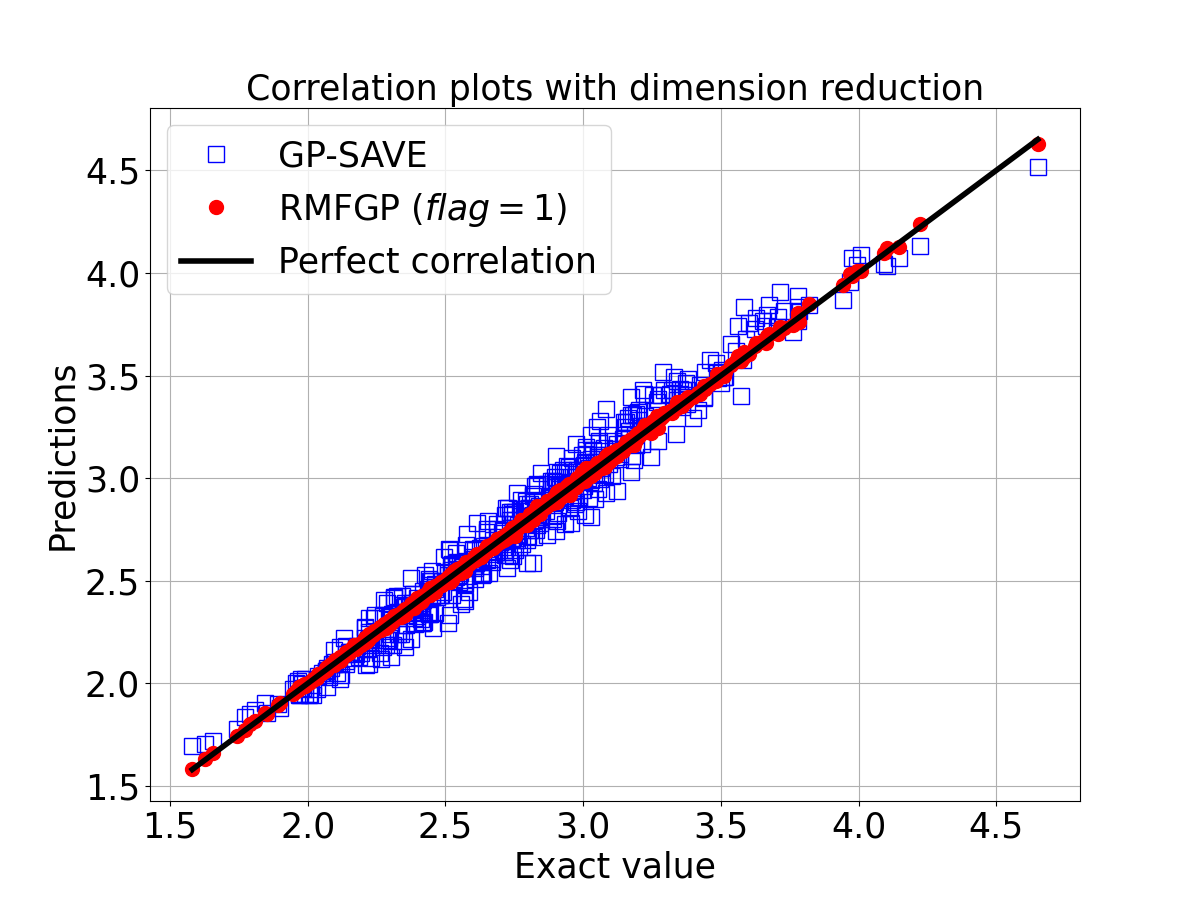}
\label{fig:NonlinearCorSAVE2-20}}

\caption{Correlation plots of nonlinear examples at $N_H=20$: (a) RMFGP ($flag=0$) vs. GP; (b) RMFGP ($flag=1$) vs. GP-SAVE. For all figures, axis x is the exact values and axis y is the predictions at the test points. The black solid line is the perfect correlation.}
\label{fig:NonlinearCor}
\end{figure}

Figure $6$ is the prediction plot at $N_H=20$. The black star line is the exact prediction with $x_d$ computed by the true dimension reduction matrix. The red circle line obtained by RMFGP ($flag=1$) fits the curve well. The blue square line has a large error at some locations of $x_d$. The successful estimation of central subspace as well as the predictions on test set proves the ability of our method to exclude the effect of noises with relatively small set of highly accurate data, which is useful in many real world applications where the high-fidelity data is expensive or hard to collect.

\begin{figure}[h]
\centering
\includegraphics[width=0.6\textwidth]{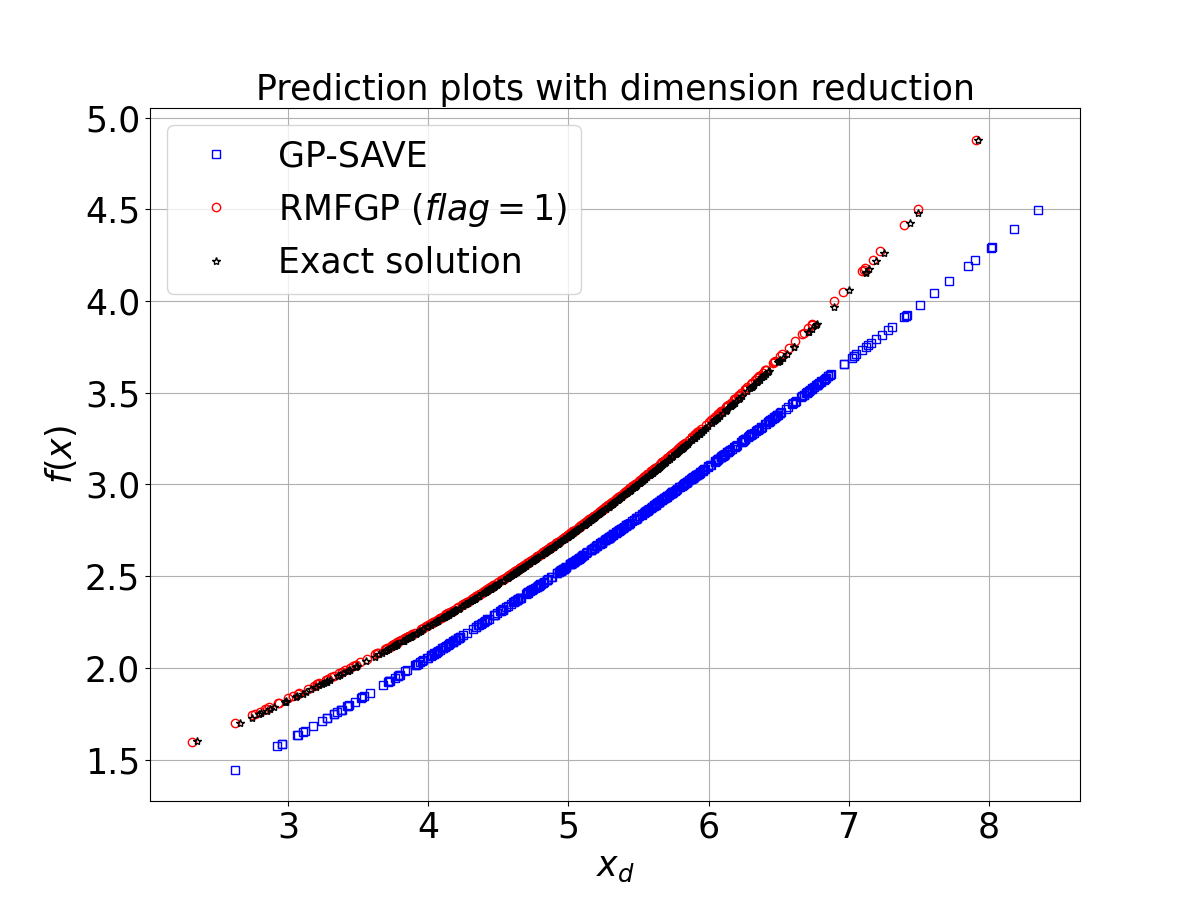}
\label{fig:NonlinearplotSIR-20}
\caption{Prediction plots of nonlinear example at $N_H=20$: RMFGP ($flag=1$) vs. GP-SAVE. The axis x is the test inputs after dimension reduction represented by $x_d$ and the axis y is the corresponding observations. The number of original dimensions is $p=10$. The number of training data is $N_L = 200$ and $N_H = 20$.}
\label{fig:nonlinearplot}
\end{figure}

\subsection{Advection equation}
This example is aimed to exam the performance of RMFGP in the stochastic partial differential equation problems. Consider the one dimensional differential equation:
\begin{equation}
    \frac{\partial}{\partial t} u(x, t; \xi)+ \frac{a}{4} \sum_{i=1}^5 \bold{\xi}_i \frac{\partial}{\partial x} u(x, t; \xi) = 0 
\end{equation}
with the initial condition:
$$ u(x, 0; \xi) = sin(\pi (x + 1)) + 1 $$
where $a$ is a constant coefficient, $x \in [0,1]$ and $\bold{\xi} = (\xi_1, \cdots, \xi_5) \in [0,1]^5$ is a random vector.
Under this setting we have analytical solution for this equation denoted as $u_H$:
$$ u_H(x) = sin(\pi (x - \frac{a}{4} t \sum_{i=1}^5 \bold{\xi}_i + 1)) + 1 $$
The low-fidelity data is sampled from the following function:
$$ u_L(x) =  sin(\pi (x - \frac{a}{4} t \sum_{i=3}^5 \bold{\xi}_i + 1)) + 1 $$
The input random vector $\xi$ is generated by i.i.d uniformly distribution in $[0,1]^5$ and the constant $a$ is fixed to be $1$. Compared to the previous two examples, there is some missing information in $u_L$ in this example. The high and low-fidelity function values are computed at $x=0.5$ and $t=1$. The true reduced dimension is $d=1$ from the analytical solution of the equation. The true reduction matrix is $A = span\{\beta_1\}$, where $\beta_1 = (1,1,1,1,1)$. Table $7$ shows the $G(k)$ values computed by BIC method. It reaches maximum at $k=1$, which indicates the estimated reduced dimension is $\hat{d}=1$. In the numerical experiment, the number of low-fidelity training data is set to be $N_L=200$ and the number of test data is $N_T = 500$. For all four cases with different number of $N_H$, the experiment starts at $(N_H-10)$ and the active learning process add $5$ samples per iteration with $2$ iterations before stopping.

\begin{table}[h!]
\centering
\begin{tabular}{||c c c c c c||} 
 \hline
              &  k=1         &  k=2                    &   k=3                   & k=4        & k=5       \\ [0.5ex] 
 \hline
  G(k)        &  $\textbf{0.8753940}$ &   0.8464151    &    0.8165653  &  0.7861532   &  0.7552568       \\  
 \hline
\end{tabular}
\caption{BIC: G(k) for advection equation - Equation (12)}
\label{table:7}
\end{table}

Table $8$ summarize the distance defined in (9). As expected, both methods demonstrate the consistency but the proposed RMFGP($flag=1$) method outperforms GP-SAVE on all sample sizes. Note that when $N_H$ is not sufficient, GP-SAVE can not detect central subspace well. The relative error in Table $9$ and the MSE plot in Figure $7$ confirms this conclusion. RMFGP has a smaller error and converge faster compared to GP. The correlation plot in Figure $8$ and the prediction plot in Figure $9$ shows that RMFGP achieves a smaller generalization error and regresses the curve better than GP with SAVE method. This example indicates the capability of RMFGP to successfully approximate the central subspace in a stochastic differential equation problem with missing information in low-fidelity model $u_L$. 

\begin{table}[h!]
\centering
\begin{tabular}{||c c c c c||} 
 \hline
              &  $N_H=20$      &  $N_H=25$      &    $N_H=30$   & $N_H=35$ \\ [0.5ex] 
 \hline
 RMFGP ($flag=1$)  & 0.249067      &    0.114606    &    0.072898   & 0.066792     \\ 
 \hline
 GP-SAVE     &  1.410894     &   1.269664   &   0.483262  &   0.450924  \\  
 \hline
\end{tabular}
\caption{Error of the dimension reduction matrix measured by the metric $m(A,\hat{A})$ with different number of high-fidelity samples for advection equation - Equation (12). In this experiment, constant $a$ is set to be 1 and $x=0.5$, $t=1$ are fixed. The number of low-fidelity samples is fixed to be $N_L = 200$. There are two iterations to add high-fidelity samples in Bayesian active learning process with 5 points added per iteration. For RMFGP method, the number of dimensions of the inputs is first reduced to dimension $s = 3$ from the number of original dimensions $p=5$. Then Gaussian process dimension reduction technique is applied to reduce the dimension from $s=3$ to $d=1$. The additional number of hyper-parameters needed to optimized in this step is $3$.}
\label{table:8}
\end{table}

\begin{table}[h!]
\centering
\begin{tabular}{||c c c c c||} 
 \hline
                     & $N_H= 20$   &   $N_H=25$     &    $N_H=30$     &    $N_H=35$ \\ [0.5ex] 
 \hline
RMFGP ($flag=0$) & 0.645469   &   0.084900   &   0.068312  &    0.061210 \\ 
 \hline
 GP                & 1.025144   &   0.964663   &   0.705019    &  0.672353\\
 \hline
RMFGP ($flag=1$) & 0.277659 & 0.260726    &   0.216323 &    0.084661 \\ 
 \hline
 GP-SAVE         & 0.743488 & 0.665575  &   0.406130   &   0.384018  \\ 
\hline
\end{tabular}
\caption{Relative error $e$ of RMFGP model compared to standard GP for advection equation - Equation (12). If $flag=0$, the inputs are simply rotated by the rotation matrix from RMFGP model before fed into a new GP surrogate model. It is compared to a standard GP model. If $flag=1$, the inputs are reduced to dimension $d=1$. For comparison, the inputs for the standard GP are reduced to dimension $d=1$ by a reduction matrix computed by SAVE method using the same number of high-fidelity training points.}
\label{table:9}
\end{table}

\begin{figure}[h]
\centering
\subfloat[a][]{
\includegraphics[width=0.45\textwidth]{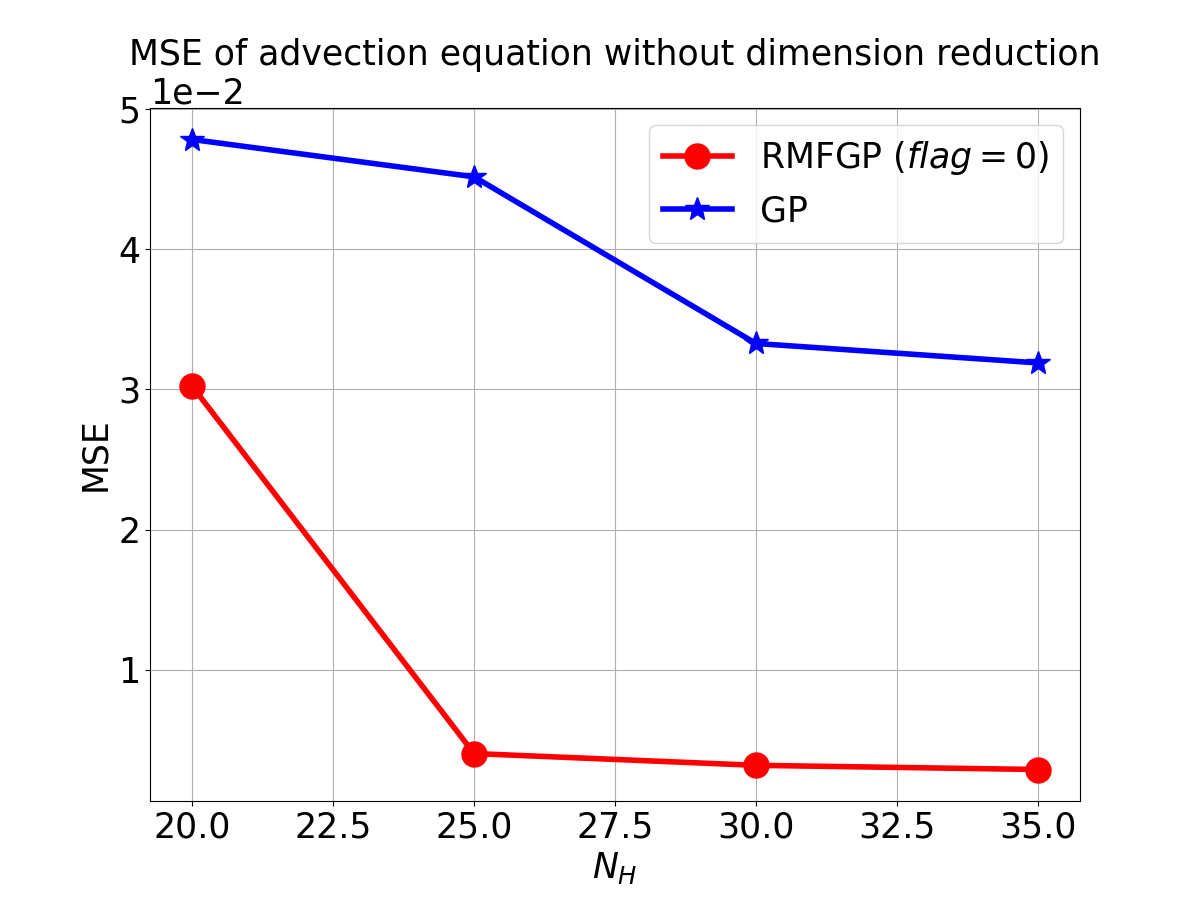}
\label{fig:MSE_adv1}}
\qquad
\subfloat[b][]{
\includegraphics[width=0.45\textwidth]{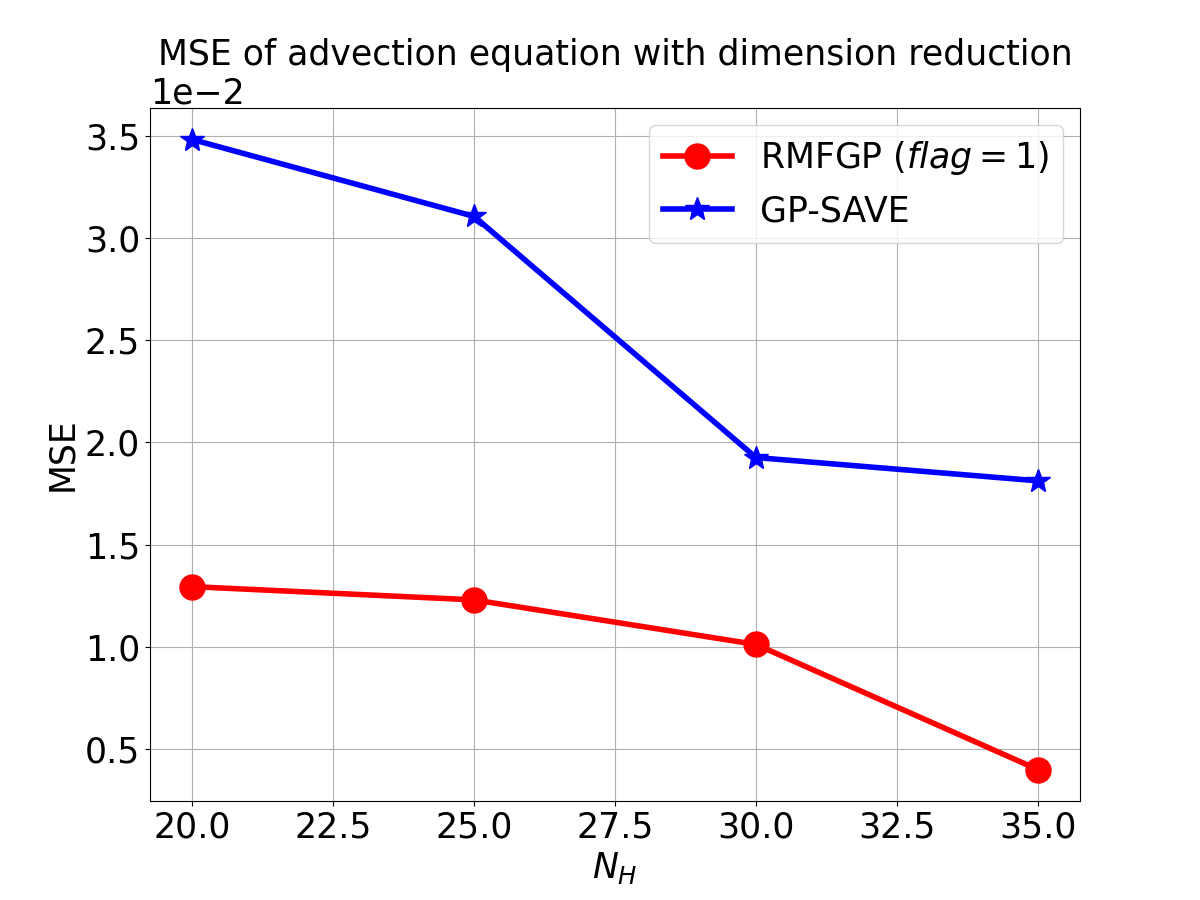}
\label{fig:MSE_adv2}}
\caption{MSE of advection equation - Equation (12): (a) Models without dimension reduction: RMFGP ($flag=0$) vs. GP; (b) Models with dimension reduction: RMFGP ($flag=1$) vs. GP-SAVE. For both (a) and (b), axis x is the number of high-fidelity samples $N_H$ and the number of low-fidelity sample $N_L$ is fixed to be 200. The number of original dimensions is $p = 5$. Constant $a=1$, $x=0.5$ and $t=1$ are fixed.}
\label{fig:MSE_adv}
\end{figure}

\begin{figure}[h]
\centering
\subfloat[a][]{
\includegraphics[width=0.45\textwidth]{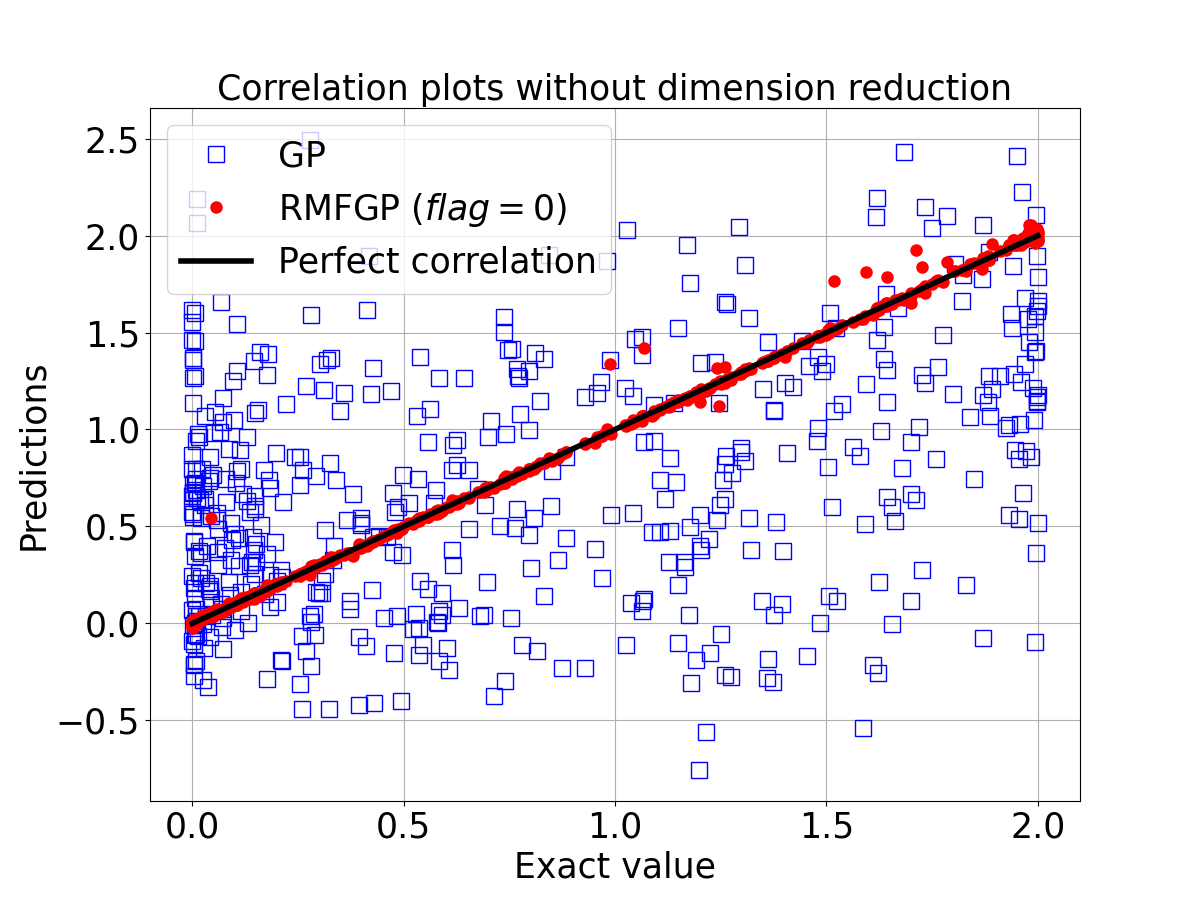}
\label{fig:AdvCorSAVE1-30}}
\qquad
\subfloat[b][]{
\includegraphics[width=0.45\textwidth]{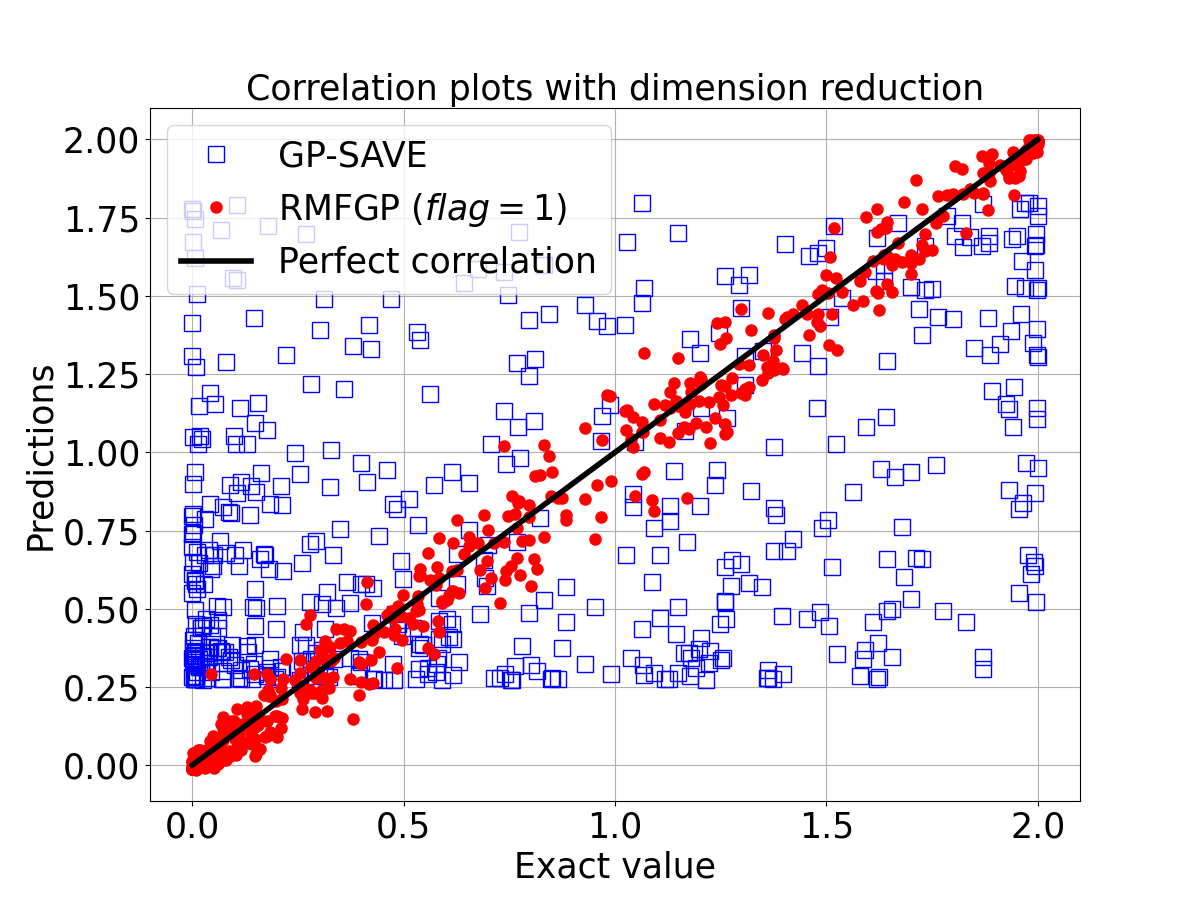}
\label{fig:AdvCorSAVE2-30}}

\caption{Correlation plots of advection equation - Equation (12) at $N_H = 30$: (a) RMFGP ($flag=0$) vs. GP; (b) RMFGP ($flag=1$) vs. GP-SAVE. For all figures, axis x is the exact values and axis y is the predictions at the test points. The black solid line is the perfect correlation.
}
\label{fig:AdvCor}
\end{figure}

\begin{figure}[h]
\centering
\includegraphics[width=0.6\textwidth]{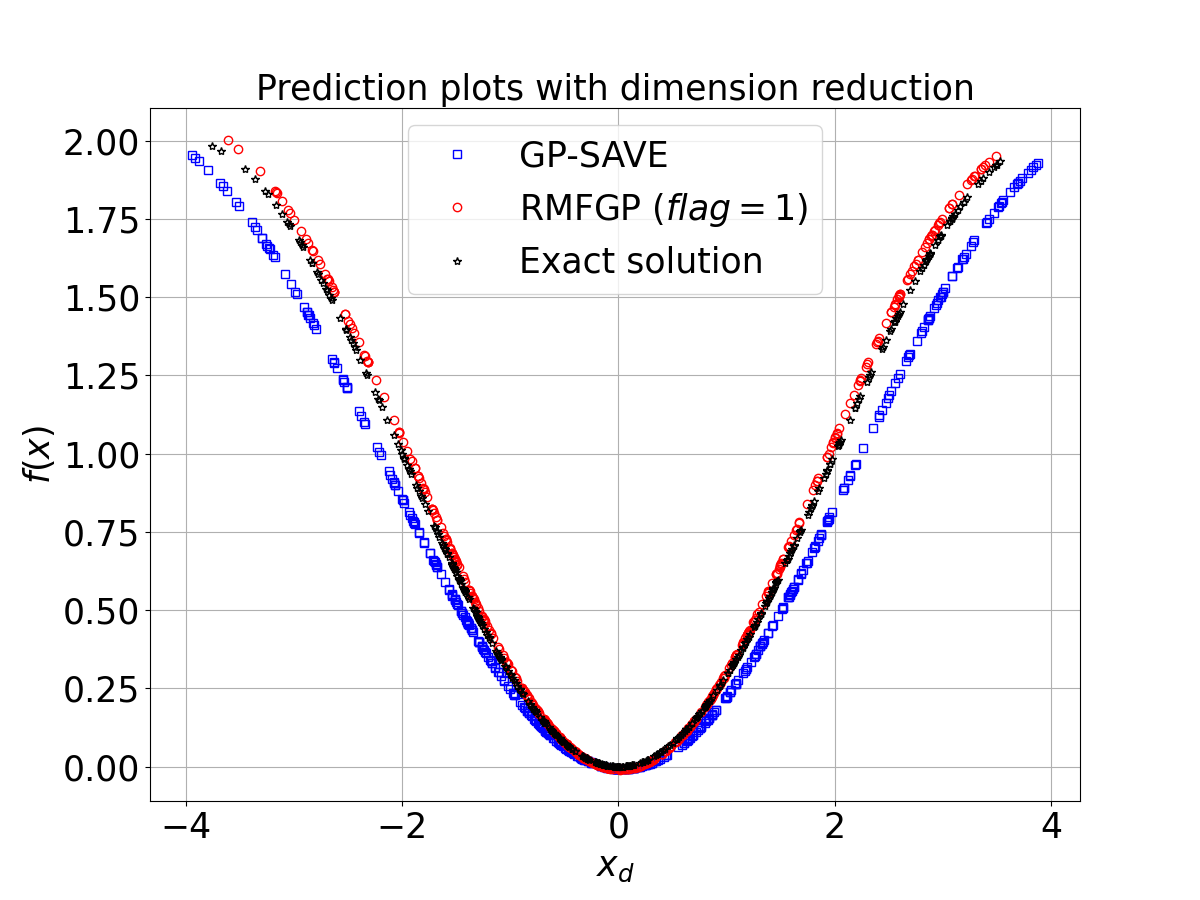}
\label{fig:AdvplotSAVE-30}

\caption{Prediction plots of advection equation - Equation (12) at $N_H = 30$: RMFGP ($flag=1$) vs. GP-SAVE. The axis x is the test inputs after dimension reduction represented by $x_d$ and the axis y is the corresponding observations. The number of original dimensions is $p=5$. The number of training data is $N_L = 200$ and $N_H = 30$.}
\label{fig:Adv prediction}
\end{figure}

The proposed method gives us a dimension reduction matrix $M$ if $flag=1$. Then, we can build a new Gaussian process surrogate with $M$ by pre-processing all training data with $M$ to reduce the input dimension to $d=1$. In order to perform an uncertainty propagation analysis for this model, we fix $t=1$ and chose $50$ evenly spaced locations in $[0,1]$ for $x$ in advection equation. Then, 2000 samples for $\bold{\xi}$ are drawn from an i.i.d uniformly distribution. Figure $10$ presents the average of means and standard deviations(std) of those 2000 cases along with $50$ different $x$ values in $[0,1]$. The ground truth is the black line. We can see our RMFGP method represented by red square line outperforms the comparison GP method represented by blue star line for both mean and std values. The green diamond line is obtained by pure SAVE method using large enough training data, i.e. $10000$ high-fidelity data samples, while there are only $35$ high-fidelity samples in our proposed method.  

\begin{figure}[h]
\centering
\subfloat[a][]{
\includegraphics[width=1\textwidth]{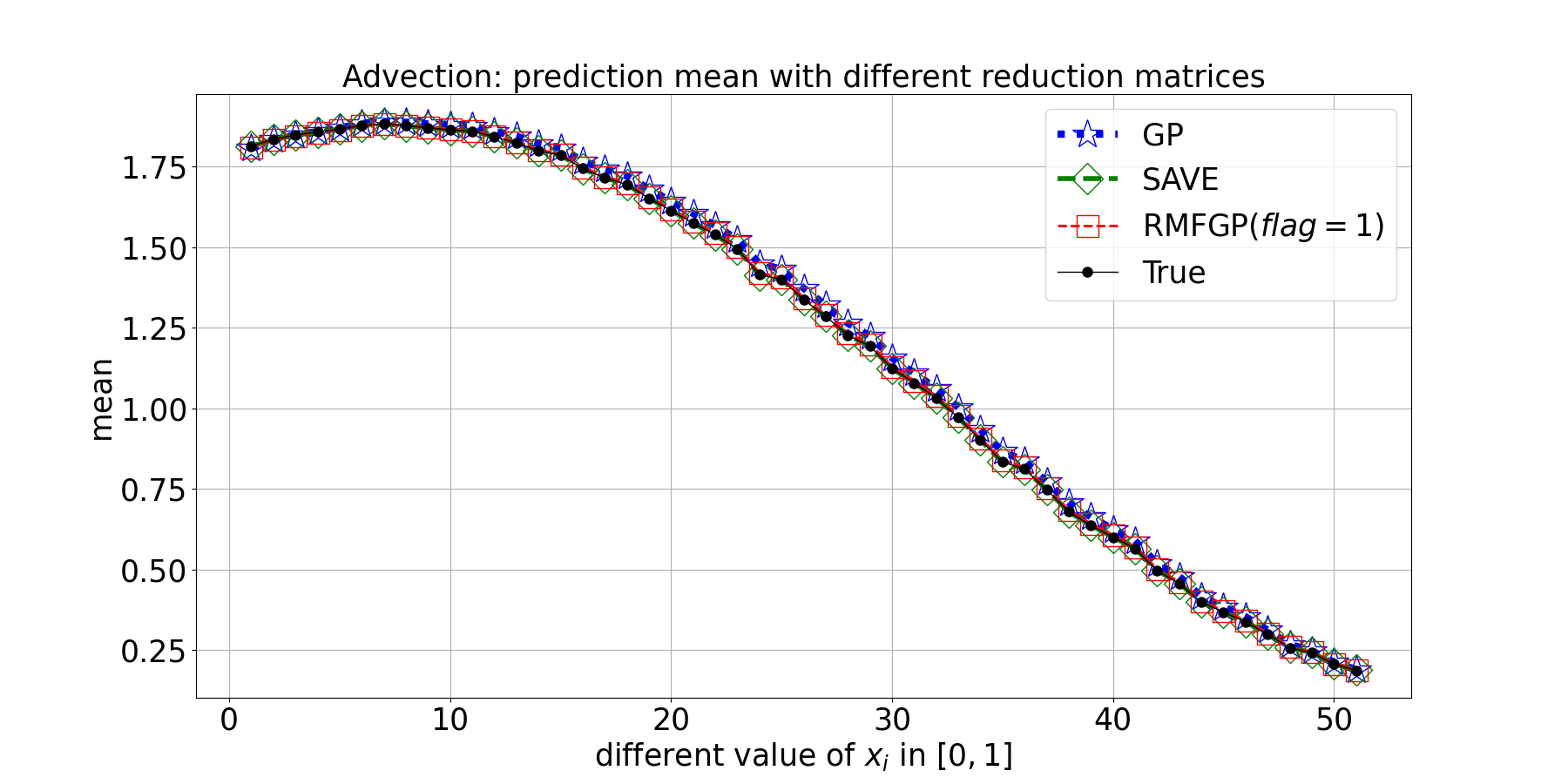}
\label{fig:AdvUQmean40}} \\
\qquad
\subfloat[b][]{
\includegraphics[width=1\textwidth]{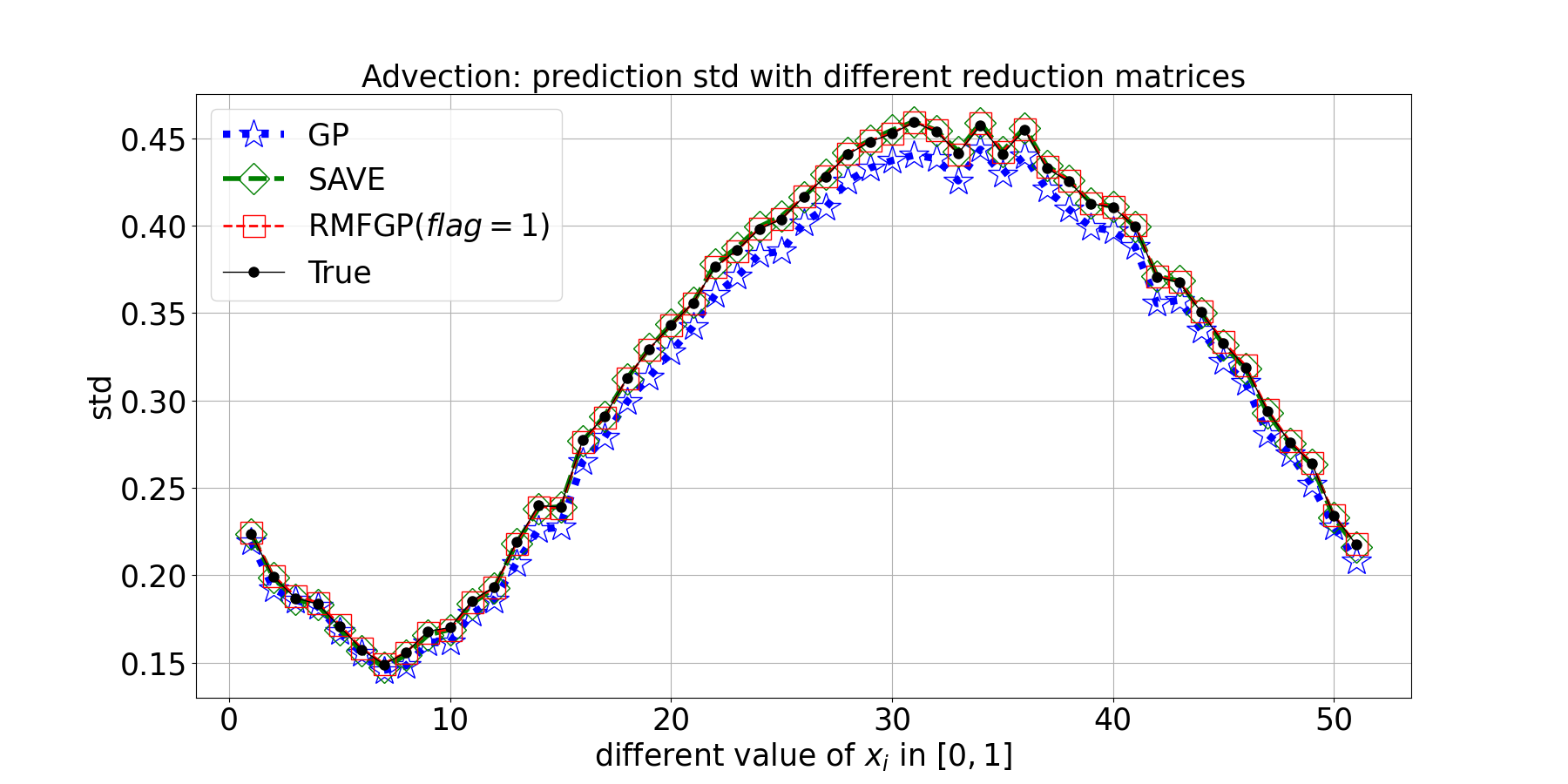}
\label{fig:AdvUQstd40}}

\caption{Average prediction means and standard deviations of advection equation - Equation (12) at $N_H = 35$: (a) average mean; (b) average standard deviation. For all figures, the axis $x$ represents the indexes of different $x$ values in $[0,1]$ and the axis $y$ represents the average mean or std values. The ground truth is the black line. The green diamond line is obtained by pure SAVE method with $10000$ high-fidelity data points. The blue star line is results of GP-SAVE method. Our RMFGP method with $flag=1$ is represented by the red square line. Both those two methods are using $35$ high-fidelity samples.
}
\label{fig:AdvUQ}
\end{figure}


\subsection{Elliptic Equation}
The last example illustrates the performance of RMFGP in a more complicated stochastic partial differential equation. Consider the one-dimensional elliptic differential equation with a random high-order coefficient:
\begin{equation}
    -\frac{d}{dx}(a(x;\xi)\frac{du(x;\xi)}{dx})=1,\quad x\in(0,1)
\end{equation}
$$u(0)=u(1)=0,$$
where $a(x;\xi)$ has the form:
$$a_H(x;\xi) = \frac{1}{(\xi_1 + sin(x(\xi_1 + \xi_2 + \xi_3 + \xi_4)) + 1)}$$
$$a_L(x;\xi) = \frac{1}{(0.1 + sin(x(\xi_1 + \xi_2 + \xi_3 + \xi_4)) + 1)}  $$
For this elliptic equation, there is a deterministic solution as:
$$u(x) = u(0) + \int_0^x \frac{a(0)u(0)' - y}{a(y)}dy$$
Applying the boundary condition $u(0) = u(1) = 0$ we have:
$$a(0)u(0)' = \frac{\int_0^1 \frac{y}{a(y)}dy}{\int_0^1 \frac{1}{a(y)}dy}$$
In this example, the input $\xi$ is generated by i.i.d uniformly distributed random variables in $[0,1]^{6}$ and $u_H$ is computed at $x=0.7$. The high-fidelity function $u_H$ and low-fidelity function $u_L$ are obtained by applying corresponding $a_H$ and $a_L$ in the deterministic solution respectively.

The integrals in the deterministic solution are computed by highly accurate numerical integrations. Unlike the previous example, there is no analytical expression for $u$. The exact central subspace can not be concluded by the expression of the equation directly. Instead, it is computed by traditional SAVE method using $10000$ samples from the true distribution. The true reduced dimension is computed through these samples by BIC and the result of $G(k)$ is shown in Table $10$. The estimated dimension turns out to be $\hat{d}=2$. In this example, the number of low-fidelity points is set to be $N_L = 200$ and the number of test points is set to be $N_T=500$. For all different cases, the start number of high-fidelity samples is $(N_H-5)$ with $2$ points added in the first iteration and $3$ points added in the second.

\begin{table}[h!]
\centering
\begin{tabular}{||c c c c c c c||} 
 \hline
              &  k=1         &  k=2                    &   k=3                   & k=4        & k=5      & k=6       \\ [0.5ex] 
 \hline
  G(k)        &  0.7018816 &   $\textbf{0.7323037}$    &    0.7094890  &  0.6859600   &  0.6621426   & 0.6379765     \\  
 \hline
\end{tabular}
\caption{BIC: G(k) for elliptic equation - Equation (13)}
\label{table:9}
\end{table}

Table $11$ summarize the accuracy of the estimated central subspace. Table $12$ shows the relative error at various high-fidelity samples sizes on two type of methods RMFGP and GP. Figure $11$ is the MSE plot for RMFGP model with $flag=0$ or $1$ compared to corresponding GP and GP-SAVE model. Figure $12$ denotes the correlation between true observations and the prediction values at $N_H=25$. All figures shows that RMFGP has a better performance in estimating both central subspace and prediction on test set in the similar manner as in previous examples.
When the number of high-fidelity data increases, the RMFGP with $flag=0$ performs the best. This is because the identification of principle directions can improve the prediction performance. The difference between rotated model with $flag=0$ and reduced model wit $flag=1$ will reduce with the increase of the accuracy for estimated central subspace. 

\begin{table}[h!]
\centering
\begin{tabular}{||c c c c c||} 
 \hline
              &  $N_H=20$      &  $N_H=25$      &    $N_H=30$    & $N_H=35$ \\ [0.5ex] 
 \hline
 RMFGP ($flag=1$)  &  0.159129      &   0.141972  &   0.122867  &   0.120856   \\ 
 \hline
 GP-SAVE     &  0.921045   &  0.387880   &   0.250663 &   0.238877 \\  
 \hline
\end{tabular}
\caption{Error of the dimension reduction matrix measured by the metric $m(A,\hat{A})$ with different number of high-fidelity samples for elliptic equation - Equation (13). In this experiment, $x=0.7$ is fixed. The number of low-fidelity samples is fixed to be $N_L = 200$. There are two iterations to add high-fidelity samples in Bayesian active learning process with 5 points added per iteration. For RMFGP method, the number of dimensions of the inputs is first reduced to dimension $s = 3$ from the number of original dimensions $p=6$. Then Gaussian process dimension reduction technique is applied to reduce the dimension from $s=3$ to $d=2$. The additional number of hyper-parameters needed to optimized in this step is $6$.}
\label{table:11}
\end{table}

\begin{table}[h!]
\centering
\begin{tabular}{||c c c c c||} 
 \hline
                     & $N_H=20$    &   $N_H=25$     &    $N_H=30$     &    $N_H=35$ \\ [0.5ex] 
 \hline
 RMFGP ($flag=0$)   & 0.015076   &   0.007989  &  0.004557   & 0.001545 \\ 
 \hline
 GP                 & 0.060633 &   0.008441  &  0.006939  &  0.004409 \\
 \hline
 RMFGP ($flag=1$)  & 0.014057   &  0.011540    &   0.011276 &    0.010667 \\ 
 \hline
 GP-SAVE           & 0.059538  &  0.026454    &    0.020915   &    0.014822 \\ 
\hline
\end{tabular}
\caption{Relative error $e$ of RMFGP model compared to standard GP for elliptic equation - Equation (13). If $flag=0$, the inputs are simply rotated by the rotation matrix from RMFGP model before fed into a new GP surrogate model. It is compared to a standard GP model. If $flag=1$, the inputs are reduced to dimension $d=2$. For comparison, the inputs for the standard GP are reduced to dimension $d=2$ by a reduction matrix computed by SAVE method using the same number of high-fidelity training points.}
\label{table:12}
\end{table}

\begin{figure}[h]
\centering
\subfloat[a][]{
\includegraphics[width=0.45\textwidth]{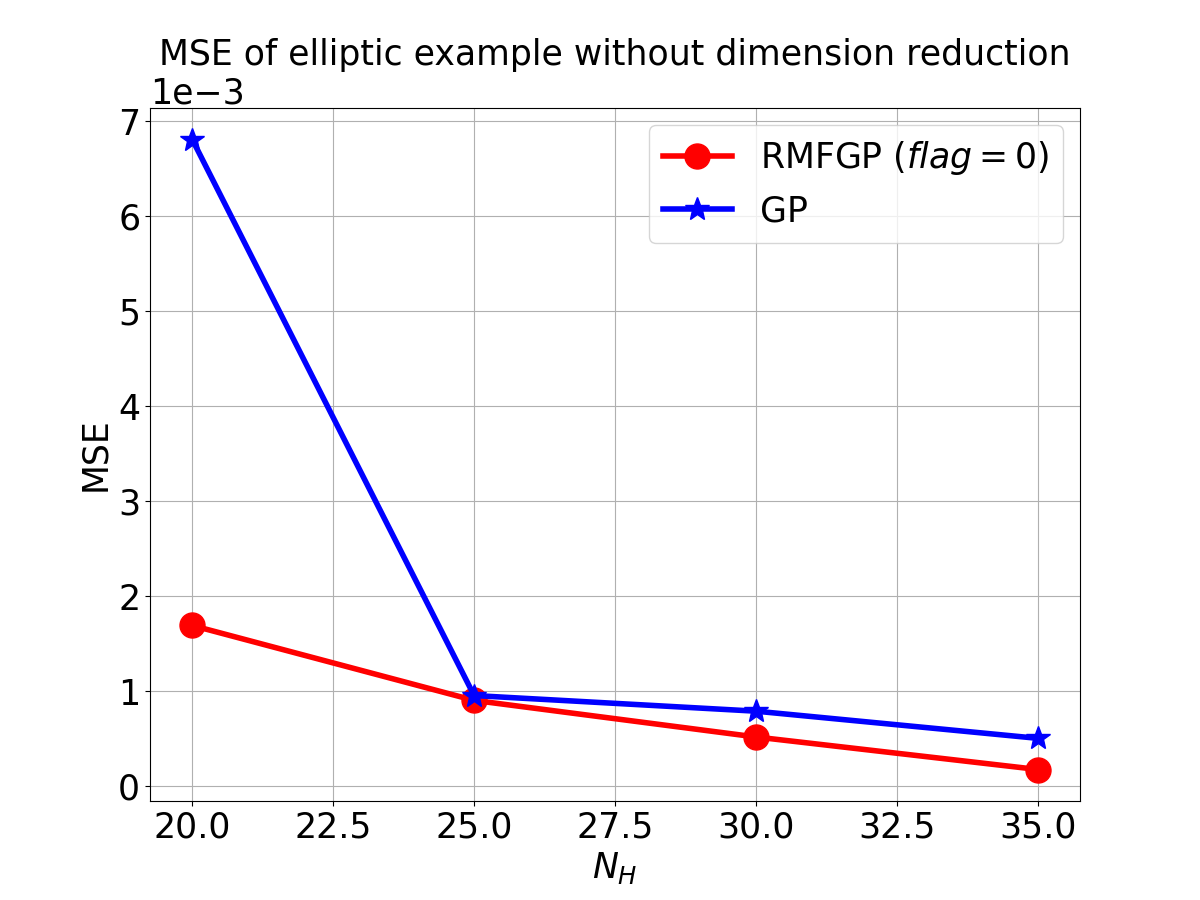}
\label{fig:MSE_elliptic1}}
\qquad
\subfloat[b][]{
\includegraphics[width=0.45\textwidth]{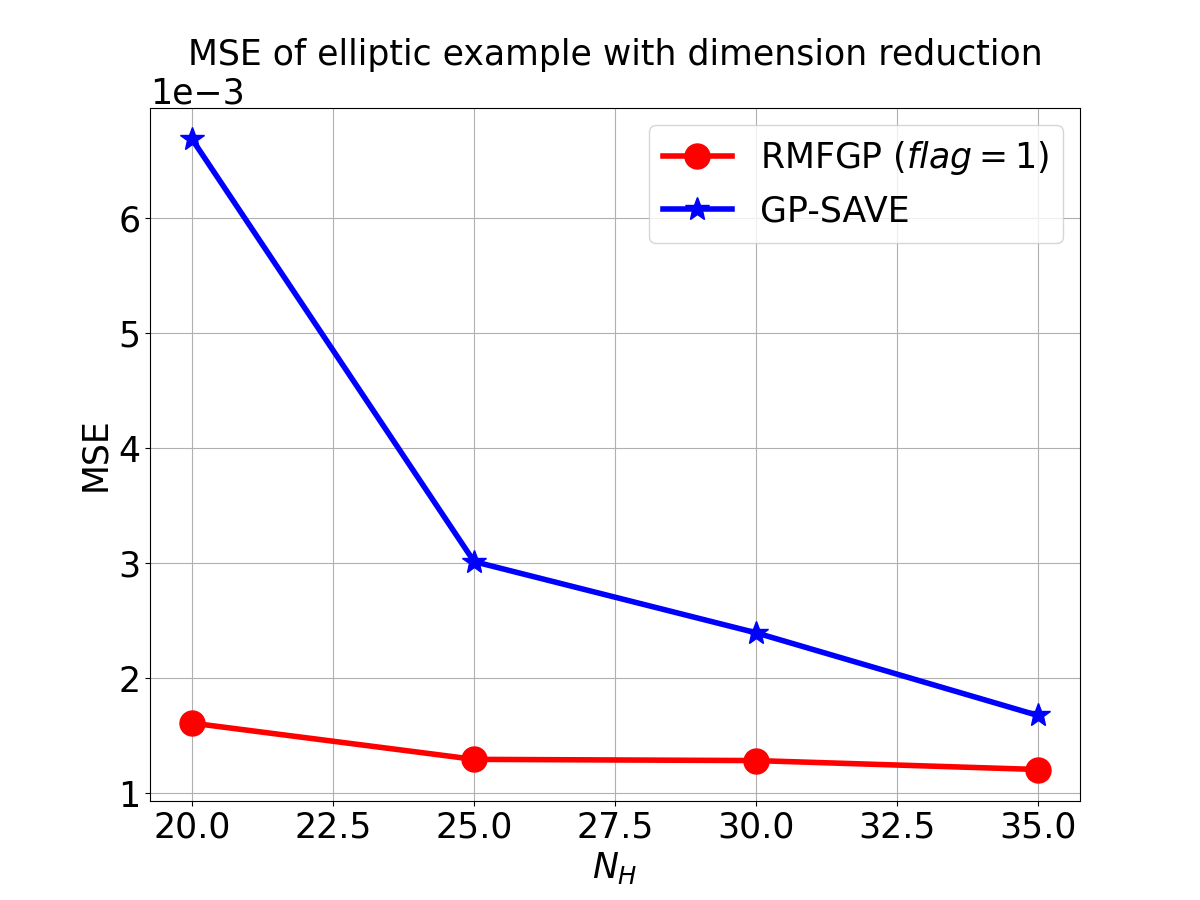}
\label{fig:MSE_elliptic2}}
\caption{MSE of elliptic equation - Equation (13): (a) Models without dimension reduction: RMFGP ($flag=0$) vs. GP; (b) Models with dimension reduction: RMFGP ($flag=1$) vs. GP-SAVE. For both (a) and (b), axis x is the number of high-fidelity samples $N_H$ and the number of low-fidelity sample $N_L$ is fixed to be $200$. The number of original dimensions is $p = 6$. }
\label{fig:MSE_elliptic}
\end{figure}

\begin{figure}[h]
\centering
\subfloat[a][]{
\includegraphics[width=0.45\textwidth]{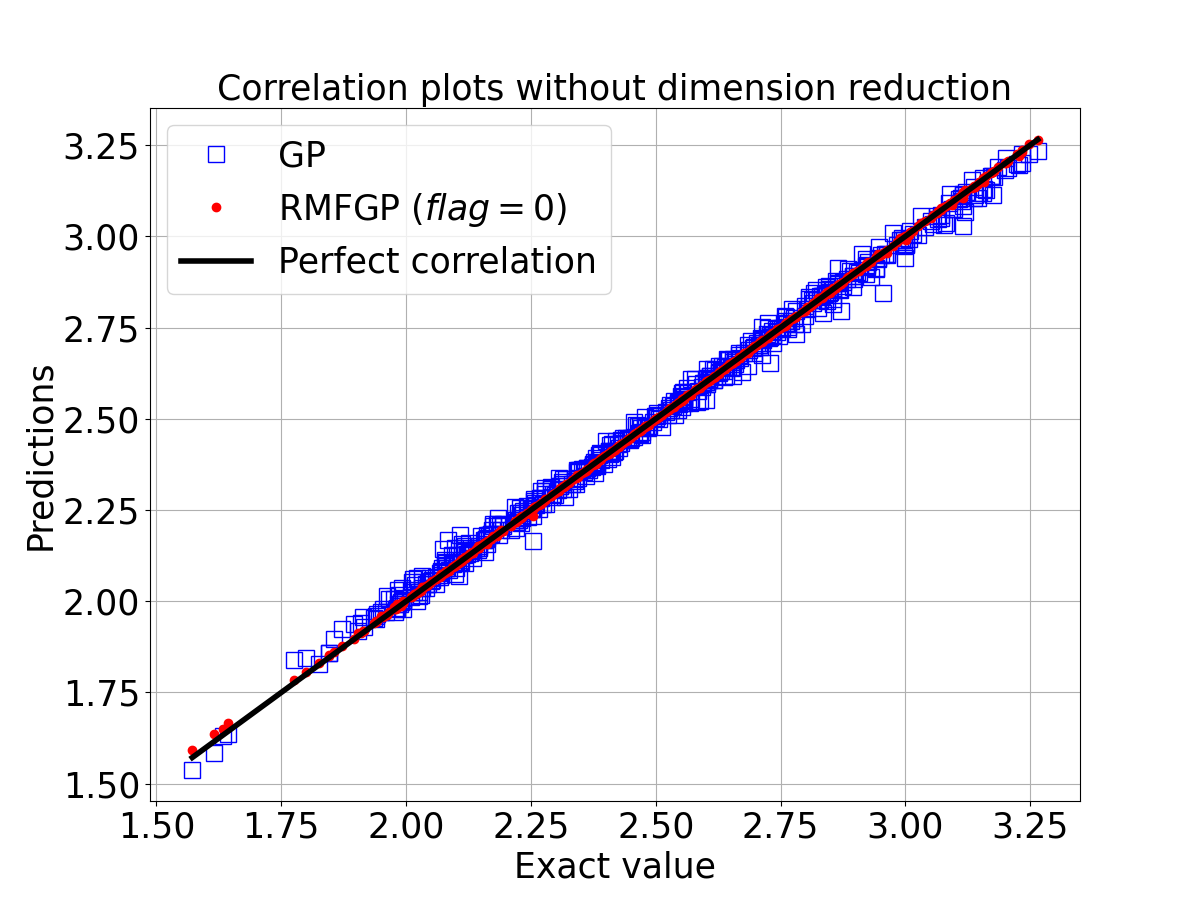}
\label{fig:AdvCorSAVE1-15}}
\qquad
\subfloat[b][]{
\includegraphics[width=0.45\textwidth]{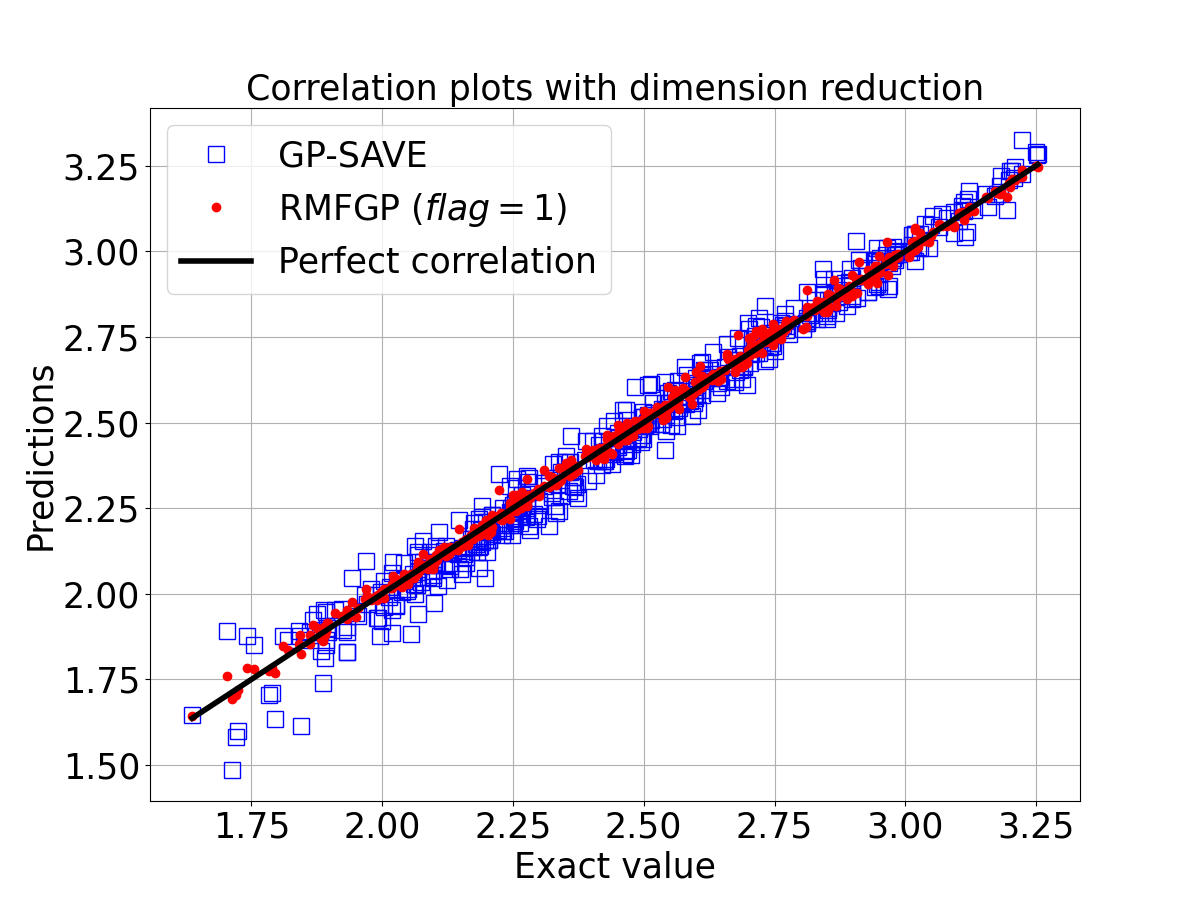}
\label{fig:EllipticCorSAVE-25}}

\caption{Correlation plots of elliptic equation - Equation (13) at $N_H = 25$: (a) RMFGP ($flag=0$) vs. GP; (b) RMFGP ($flag=1$) vs. GP-SAVE. For all figures, axis x is the exact values and axis y is the predictions at the test points. The black solid line is the perfect correlation.
}
\label{fig:EllipticCor}
\end{figure}

The uncertainty propagation analysis is also performed for this model. With the dimension reduction matrix $M$ obtained by RMFGP model with $flag=1$, a new Gaussian process surrogate can be built by pre-processing all training data with $M$ to reduce the input dimension to $d=2$. Then, 2000 samples for $\bold{\xi}$ are drawn from an i.i.d uniformly distribution. Figure $13$ presents the average of means and standard deviations(std) of those $2000$ cases along with $50$ different $x$ in $[0,1]$. The $x$-axis represents the indexes of different $x$ values. The $y$-axis is the average mean for $(a)$ and std for $(b)$ in each case. The ground truth is the black line. The green diamond line is obtained by pure SAVE method if we give large enough training data, i.e. high-fidelity data samples. Here, we give $10000$ samples in order to get this results. Note that there are only $35$ high-fidelity samples in RMFGP model. From Figure $13(a)$, all four method have similar performance. However, Figure $13(b)$ shows that our RMFGP method has smaller std compared to GP-SAVE method. This shows RMFGP is more confident about the predictions.

\begin{figure}[h]
\centering
\subfloat[a][]{
\includegraphics[width=1\textwidth]{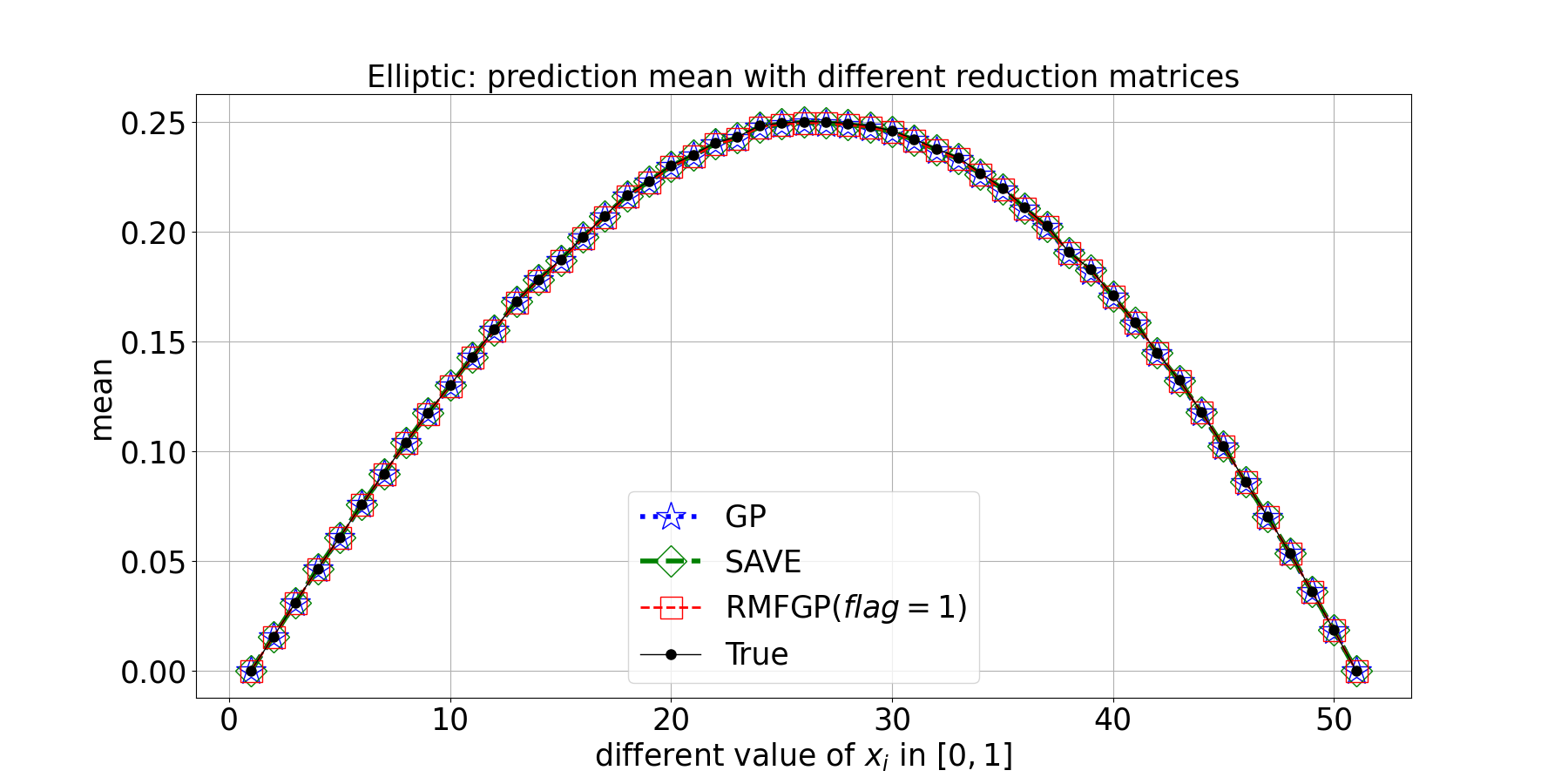}
\label{fig:EllipticUQmean40}} \\
\qquad
\subfloat[b][]{
\includegraphics[width=1\textwidth]{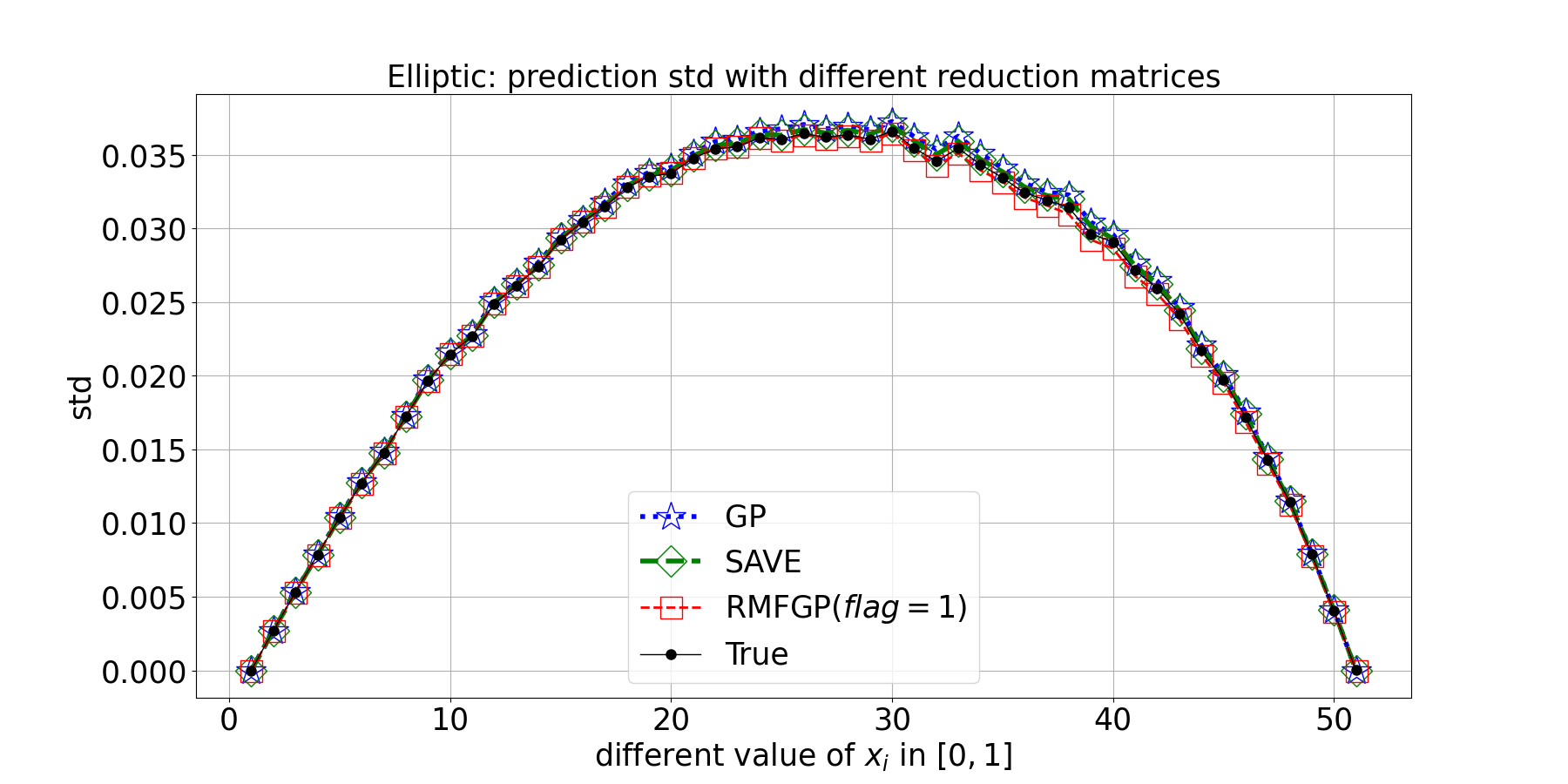}
\label{fig:EllipticUQstd40}}

\caption{Average prediction means and standard deviations of elliptic equation - Equation (13) at $N_H=35$: (a) average mean; (b) average standard deviation. For all figures, the axis $x$ represents the indexes of different $x$ values in $[0,1]$ and the axis $y$ represents the average mean or std values. The ground truth is the black line. The green diamond line is obtained by pure SAVE method with $10000$ high-fidelity data samples. The blue star line is results of GP-SAVE method. Our RMFGP method with $flag=1$ is represented by the red square line. Both those two methods are using $35$ high-fidelity samples.
}
\label{fig:EllipticUQ}
\end{figure}

\section{Conclusion}
In this paper, a new dimension reduction framework based on the multi-fidelity Gaussian process, the SAVE dimension reduction method and Gaussian process dimension reduction technique is established to estimate the central subspace and increase the prediction accuracy under the condition where only limited precise data is available. Two different approaches to build a final surrogate model can be chosen based on the $flag$ parameter in the algorithm. Based on the property that the uncertainties can be naturally quantified in a Gaussian process regression, Bayesian active learning is involved to enhance the efficiency of the method. Four numerical examples are presented in order to illustrate the ability of the proposed RMFGP model to extract the principle directions and build a corresponding surrogate model to increase the prediction accuracy under different situations.

The dimension reduction methods to computed the rotated matrices in this paper are SAVE type of methods. Other methods such as SIR, active subspace\cite{acs} can also be fed to the algorithm based on different tasks. In particular, investigating the regression tasks with missing data or labels can be interesting in the future works. Another potential future work is to find an approach to determined the optimal dimension $s$ if parameter $flag=1$ in the algorithm. 
With the proposed RMFGP model, one can build up an accurate surrogate model with lower dimensional inputs than the original data. As shown in the numerical examples, This model can be used to exclude the effect of various noises. It can also help some applications where only few indexes are allowed to represent the system. Additionally, this model requires fewer precise data to construct in a high-dimensional problem, which can save up computational resources in many applications.



\bibliographystyle{abbrv}
\bibliography{mybib}
\nocite{SPCSAD, DAS, MDRVAS, LbSDR, DDSIR, DRSLRKHS, SDRVBMM, ODRGF, ADEDRS}

\end{document}